\documentclass{article}
\usepackage{amsmath}
\usepackage{caption}

\usepackage{arxiv}

\usepackage[utf8]{inputenc} 
\usepackage[T1]{fontenc}    
\usepackage{hyperref}       
\usepackage{url}            

\usepackage{booktabs}       
\usepackage{amsfonts}       
\usepackage{nicefrac}       
\usepackage{microtype}      
\usepackage{lipsum}		
\usepackage{graphicx}
\usepackage{xcolor}
\usepackage{colortbl}

\usepackage{threeparttable}

\usepackage[frozencache,cachedir=_minted-template]{minted}
\usepackage[numbers,square, sort&compress]{natbib}
\usepackage{doi}

\usepackage{amssymb}
\usepackage{pifont}
\newcommand{\cmark}{\ding{51}}%
\newcommand{\xmark}{\ding{55}}%

\title{SurGen: 1020 H\&E-stained Whole Slide Images With Survival and Genetic Markers}

\date{} 					

\author{ \href{https://orcid.org/0000-0002-2701-3149}{\includegraphics[scale=0.06]{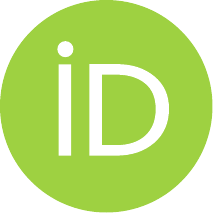}\hspace{1mm}Craig Myles}\thanks{Corresponding Author.\\Paper under review.} \\
	School of Computer Science\\
        University of St Andrews\\
	\texttt{cggm1@st-andrews.ac.uk} \\
	\And
	\href{https://orcid.org/0000-0001-9999-4292}{\includegraphics[scale=0.06]{orcid.pdf}\hspace{1mm}In Hwa Um} \\
	School of Medicine\\
	University of St Andrews\\
	\texttt{ihu@st-andrews.ac.uk} \\
        \And
        Craig Marshall \\
	Lothian Biorepository\\
	NHS Lothian\\
	\texttt{craig.marshall@nhs.scot} \\
        \And
	\href{https://orcid.org/0000-0002-0740-3668}{\includegraphics[scale=0.06]{orcid.pdf}\hspace{1mm}David Harris-Birtill} \\
	School of Computer Science\\
	University of St Andrews\\
	\texttt{dcchb@st-andrews.ac.uk} \\
        \And
	\href{https://orcid.org/0000-0001-9041-9988}{\includegraphics[scale=0.06]{orcid.pdf}\hspace{1mm}David J Harrison} \\
	School of Medicine\\
	University of St Andrews\\
	\texttt{djh20@st-andrews.ac.uk} \\
}

\renewcommand{\shorttitle}{SurGen: 1020 H\&E-stained Whole Slide Images With Survival and Genetic Markers}

\hypersetup{
pdftitle={SurGen: 1020 H\&E-stained Whole Slide Images With Survival and Genetic Markers},
pdfsubject={cs.CV},
pdfauthor={Craig GG.~Myles, et al.},
pdfkeywords={whole slide image (WSI), haematoxylin and eosin (H\&E) stain, mismatch repair (MMR), microsatellite instability (MSI), KRAS mutation, NRAS mutation, BRAF mutation, colorectal cancer, digital pathology, dataset},
}

\begin{document}
\maketitle

\begin{abstract}

\textbf{Background}: Cancer remains one of the leading causes of morbidity and mortality worldwide. Comprehensive datasets that combine histopathological images with genetic and survival data across various tumour sites are essential for advancing computational pathology and personalised medicine. \textbf{Results}: We present SurGen, a dataset comprising 1,020 H\&E-stained whole slide images (WSIs) from 843 colorectal cancer cases. The dataset includes detailed annotations for key genetic mutations (\textit{KRAS}, \textit{NRAS}, \textit{BRAF}) and mismatch repair status, as well as survival data for 426 cases. We illustrate SurGen's utility with a proof‑of‑concept model that predicts mismatch‑repair status directly from WSIs, achieving a test AUROC of 0.8273. These preliminary results underscore the dataset's potential to facilitate research in biomarker discovery, prognostic modelling, and advanced machine learning applications in colorectal cancer and beyond. \textbf{Conclusions}: SurGen offers a valuable resource for the scientific community, enabling studies that require high-quality WSIs linked with comprehensive clinical and genetic information on colorectal cancer. Our initial findings affirm the dataset's capacity to advance diagnostic precision and foster the development of personalised treatment strategies in colorectal oncology. Data available online: \href{https://doi.org/10.6019/S-BIAD1285}{https://doi.org/10.6019/S-BIAD1285}.

\end{abstract}

\keywords{whole slide image (WSI) \and haematoxylin and eosin (H\&E) stain \and mismatch repair (MMR) \and microsatellite instability (MSI) \and KRAS mutation \and NRAS mutation \and BRAF mutation \and colorectal cancer \and digital pathology \and dataset}

\section{Background}
\label{sec:background}

Colorectal cancer is among the most common and lethal cancers worldwide with over 900,000 deaths occurring each year \citep{sung2021global, bray2024global}. Advances in computational pathology and machine learning have the potential to revolutionise cancer diagnosis and treatment by enabling the analysis of complex histopathological and genetic data across various tumour types \citep{bera2019artificial, niazi2019digital}.

High-quality datasets that combine whole slide images (WSIs) with detailed clinical and genetic annotations are crucial for developing and validating computational models. However, the field currently faces significant limitations due to the scarcity of publicly available annotated datasets that integrate both imaging and non-imaging patient data \citep{abels2019computational}. Existing datasets often focus on specific cancer sites, such as breast \citep{litjens20181399, spanhol2015dataset, national2020brca}, gastric and colorectal \citep{da2022digestpath, kim2023paip}, and lung \citep{national2018luad}, or lack comprehensive annotations necessary for advanced computational pathology research. Additionally, the quality of publicly available samples can be highly variable, potentially hindering the development of robust and generalisable models \citep{abels2019computational}. The SurGen dataset addresses these gaps by providing a diverse and high-quality collection of WSIs linked with genetic mutations, mismatch repair status, and cancer staging across colorectal and neighbouring sites. Additionally, it includes survival data specifically for the primary colorectal cancer cohort, enhancing its value for prognostic studies in this prevalent cancer.

This article reports on the composition, collection, and potential applications of the SurGen dataset, highlighting its utility for both focused studies specific to primary colorectal cancer and broader investigations into metastatic tumour sites. This is particularly pertinent given that up to 50\% of patients with localised disease eventually develop metastases \citep{ciardiello2022clinical}.

The SurGen dataset is a comprehensive digital pathology resource designed to support a wide range of cancer and computational pathology research initiatives. It consists of whole slide images (WSIs) coupled with detailed clinical and genetic data, spanning colorectal regions as well as neighbouring metastatic sites. See table \ref{tab:tumour-site-overview} for a breakdown of tumour sites across the SurGen dataset. The dataset is divided into two distinct subsets:

\begin{enumerate}
    \item \textbf{SR386} (\textbf{Colorectal Cohort with Survival Data}) focuses on primary colorectal cancer, consisting of 427 WSIs from 427 cases with a focus on colorectal tumour sites. This subset includes survival data in addition to biomarker labels, such as mutation status in the \textit{KRAS}, \textit{NRAS}, and \textit{BRAF} genes, as well as mismatch repair (MMR) status. This makes it particularly valuable for research aimed at understanding the genetic and biomarker properties of colorectal cancer for the exploration and prediction of its clinical outcomes.
    
    \item \textbf{SR1482} (\textbf{Colorectal Cancer with Metastatic Sites}) is a subset that contains 593 WSIs from 416 colorectal cancer cases. This cohort includes WSIs from both primary colorectal tumours and metastatic lesions in sites such as the liver, lung, peritoneum, and others. While it does not include survival data, it offers extensive biomarker information, making it valuable for studies on genetic and molecular characteristics of colorectal cancer and its metastatic behaviour. 
\end{enumerate}

The SurGen dataset aims to facilitate research in oncology and digital pathology by providing high-quality, labelled WSIs that can be used for training and validating computational models, investigating tumour and oncological properties, and exploring biomarker-driven stratification in colorectal cancer. This article reports on the composition, collection, and potential applications of the SurGen dataset, highlighting its utility for both focused studies on colorectal cancer and broader investigations into neighbouring metastatic sites and generalised oncological understanding.

To highlight the comprehensive nature of the SurGen dataset, we compare it with several publicly available colorectal cancer datasets. Table \ref{tab:open-source-overview} summarises key attributes such as the inclusion of genetic markers, survival data, and tumour staging.

\begin{table*}[bt!]
\caption{Comparative overview of publicly available formalin-fixed-paraffin-embedded (FFPE) H\&E stained colorectal whole slide image datasets with relevant biomarker labels.}
\label{tab:open-source-overview}
\resizebox{\textwidth}{!}{%
\begin{tabular}{l|lcrrcrcccccccc}
\toprule
Dataset & Access & Origin& Cases & WSIs & Magnification  &MPP& KRAS & NRAS & BRAF & MSI/MMR & Survival & Staging & Pathological & Segmentation \\
\midrule
SurGen (Ours)                            & Public       & GBR& 843 & 1020 & 40X         &0.1112& \cmark & \cmark & \cmark & \cmark & \cmark & \cmark & \cmark &\xmark\\
PAIP \citep{kim2023paip}                 & Upon Request & KOR& 118 & 118  & 40X         &0.2522&\xmark&\xmark&\xmark& \cmark &\xmark&\xmark&\xmark& \cmark \\
TCGA-COAD \cite{weinstein2013cancer}     & Public       & USA& 451 & 459  & 20X or 40X  &*0.2436& \cmark & \cmark & \cmark & \cmark & \cmark & \cmark & \cmark &\xmark\\
TCGA-READ \cite{weinstein2013cancer}     & Public       & USA& 164 & 165  & 20X or 40X  &*0.2427& \cmark & \cmark & \cmark & \cmark & \cmark & \cmark & \cmark &\xmark\\
CPTAC-COAD \citep{cptac_2020}            & Public       & USA& 105 & 220  & 40X         &0.2501& \cmark & \cmark & \cmark & \cmark &\xmark& \cmark & \cmark &\xmark\\
CRC-Orion \citep{wala2024integrating}    & Public       & USA& 40  & 42   & 20x         &0.3250& \cmark & \cmark & \cmark & \cmark & \cmark & \cmark & \cmark & \cmark \\
\bottomrule
\end{tabular}%
}
\begin{tablenotes}
\item Note: cases are only counted if at least one diagnostic whole slide image (WSI) is available per clinical record. Note that reported case counts may differ across publications due to varying inclusion criteria and filtering methods. This table does not include any tumour microarray (TMA) or patch-based datasets. MPP = Microns per pixel. MPP values marked with * are mean values across the cohort, with ranges: TCGA-COAD (0.2325-0.2527), TCGA-READ (0.2325-0.2520). 
\end{tablenotes}
\end{table*}

Other digital‑pathology collections also exist, for instance, the PLCO Cancer Screening Trial~\citep{zhu2013prostate} offers controlled access to colorectal WSIs (approximately 2,800 images from 749 cases) but does not provide tumour level molecular annotation, while the open‑access HunCRC~\citep{pataki2022huncrc} biopsy dataset contains 200 annotated slides yet lacks molecular and survival data.

As shown in Table~\ref{tab:open-source-overview}, the SurGen dataset provides a valuable addition to publicly available resources, uniquely integrating high-resolution WSIs with detailed genetic, clinical, and survival data. While datasets such as TCGA-COAD, TCGA-READ, and CPTAC-CRC offer comprehensive genomic sequencing data, SurGen complements these resources by focusing on key colorectal cancer biomarkers (\textit{KRAS}, \textit{NRAS}, \textit{BRAF}, MSI/MMR) and survival outcomes. Moreover, its consistent high-resolution scanning at 40× magnification across all slides ensures uniform image quality, addressing variability seen in some datasets, such as TCGA.

SurGen is among the largest publicly available colorectal cancer WSI datasets, with 1,020 slides from 843 cases, exceeding the combined slide count of TCGA-COAD, TCGA-READ, and CPTAC-CRC. While SurGen's genomic annotation is focused on specific biomarkers, its scale, resolution, and inclusion of survival data make it particularly well-suited for computational pathology research, prognostic modelling, and biomarker classification in colorectal cancer.

\section{Data Description}

This section provides an overview of the SurGen dataset, which includes whole slide images (WSIs) and corresponding clinical and genetic data. The dataset is intended to support research in cancer and computational pathology, offering a resource for studying genetic mutations, mismatch repair status, and patient survival outcomes. Below is a detailed description of the data and its collection process.

\begin{figure}[bt!] 
\centering
\includegraphics[width=0.75\linewidth]{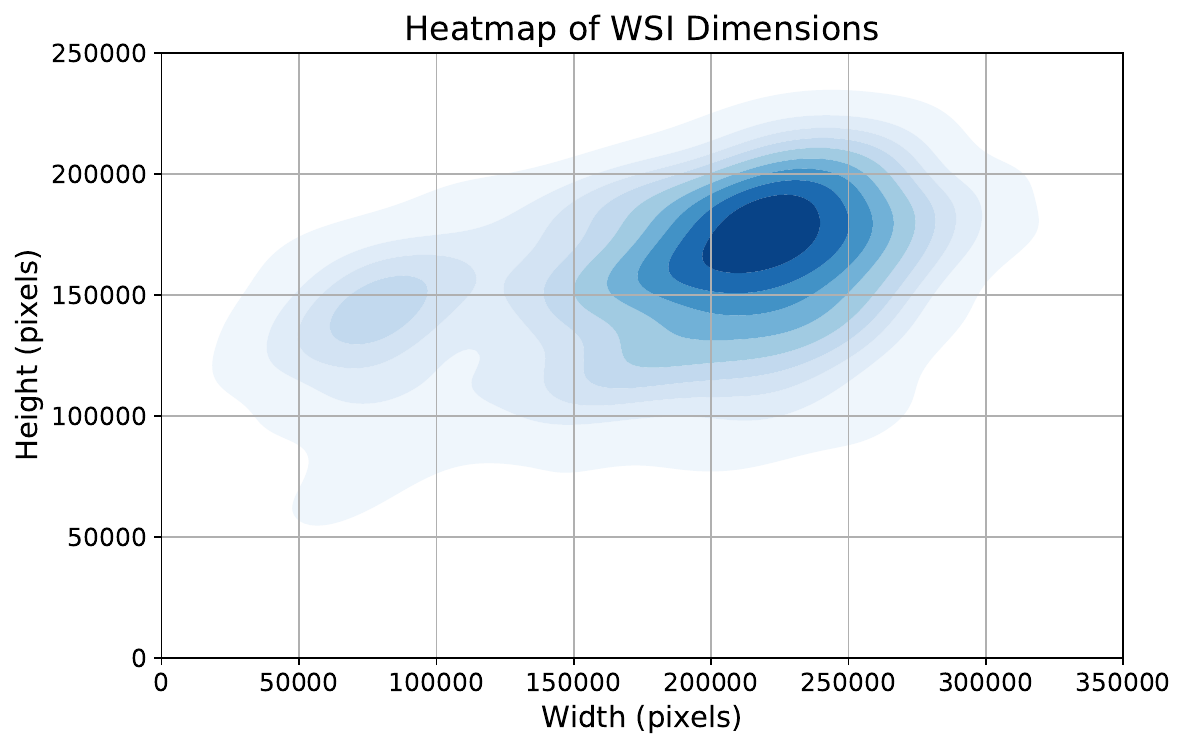}
\caption{Heatmap of WSI dimensions (in pixels) across the SurGen dataset, illustrating the variability in image sizes due to differing tissue sample areas.}\label{fig:heatmap-pixels}
\end{figure}

\begin{figure*}
\centering
\includegraphics[width=0.95\textwidth]{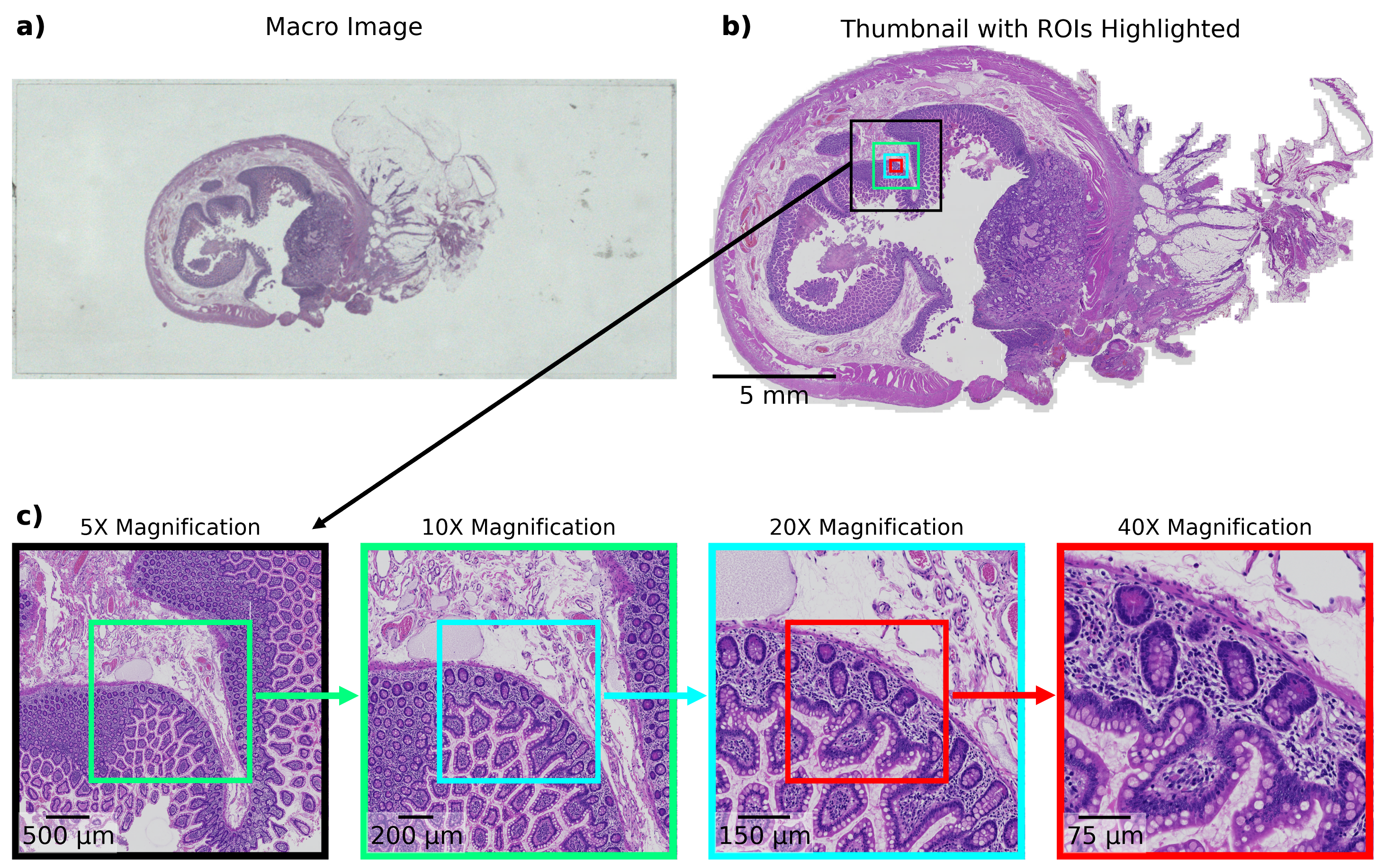}
\caption{Hierarchical zoom visualisation of case SR1482\_T412 with dimensions $242{,}506 \times 134{,}026$ pixels, corresponding to $26{,}974.20 \times 14{,}907.85 \, \mu m$. A) A low-resolution macro image of the whole slide, providing full anatomical context. B) Digitised whole slide image viewed at low-magnification. C) Successive zoom-ins of the selected region from b), providing increased granularity, enabling detailed examination of tissue structures while retaining the broader context. This hierarchical approach allows comprehensive visual exploration of tissue characteristics at varying scales. In practice, the pyramid levels are typically generated via Gaussian down-sampling to simulate various levels of magnification but enable an immediate interface for retrieving images at varying resolutions.}\label{fig:wsi-zooming}
\end{figure*}

Each WSI in the SurGen dataset is scanned at 40× (0.1112$\mu m$ per pixel) magnification, resulting in ultra-high-resolution images with pixel dimensions averaging $189{,}662\times156{,}059$ pixels. Figure \ref{fig:heatmap-pixels} illustrates the spread of WSI dimensions across the SurGen dataset. The digital pathology images are stored in the CZI file format, which supports hierarchical pyramidal data structures for efficient storage and retrieval. Figure \ref{fig:wsi-zooming} demonstrates the level of granularity accessible via the ultra-high-resolution WSIs.

\begin{table}
    \centering
\caption{Tumour Site Counts for SurGen, SR386, and SR1482}
\label{tab:tumour-site-overview}
    \begin{tabular}{lccc}
        \toprule
        Tumour Site&  SurGen&  SR386& SR1482\\
        \midrule
        Rectum&  \cellcolor{green!25}276&  \cellcolor{green!25}166& \cellcolor{green!25}110\\
        Sigmoid Colon&  \cellcolor{green!25}142&  \cellcolor{green!25}89& \cellcolor{green!25}53\\
        Caecum&  \cellcolor{green!25}118&  \cellcolor{green!25}64& \cellcolor{green!25}54\\
        Ascending Colon&  \cellcolor{green!25}99&  \cellcolor{green!25}43& \cellcolor{green!25}56\\
        Transverse Colon&  \cellcolor{green!25}46&  \cellcolor{green!25}25& \cellcolor{green!25}21\\
        Liver&  \cellcolor{green!25}38&  0& \cellcolor{green!25}38\\
        Descending Colon&  \cellcolor{green!25}33&  16& \cellcolor{green!25}17\\
        Splenic Flexure&  \cellcolor{green!25}22&  14& \cellcolor{green!25}8\\
        Hepatic Flexure&  16&  7& \cellcolor{green!25}9\\
        Peritoneum/Omentum&  16&  0&  \cellcolor{green!25}16\\
        Appendix& 9& 1&8\\
        Lung& 4& 0&4\\
        Lymph Nodes& 4& 0&4\\
        Small Bowel& 3& 0&3\\
        Bladder& 3& 0&3\\
        Gall Bladder& 2& 0&2\\
        Pelvis& 2& 0&2\\
        Site Unknown& 2& 2&0\\
        Kidney& 1& 0&1\\
        Throat/Vocal Cords& 1& 0&1\\
        Adrenal Gland& 1& 0&1\\
        Umbilical Area& 1& 0&1\\
        Spine& 1& 0&1\\
        Perineal Area& 1& 0&1\\
        Duodenum& 1& 0&1\\
        Ureter& 1& 0&1\\
        \bottomrule
    \end{tabular}

    \footnotesize{Note: Green-shaded cells indicate tumour sites within the top cumulative ranges of approximately 90\% for each respective dataset (SurGen, SR386, and SR1482). The exact highlighted cumulative totals are 91.81\%, 90.63\%, and 91.83\%, respectively. These sites collectively account for the majority of tumour occurrences in each dataset.}
    
\end{table}

\subsection{Patient Demographic}

\begin{figure}[bt!] 
\centering
\includegraphics[width=0.75\linewidth]{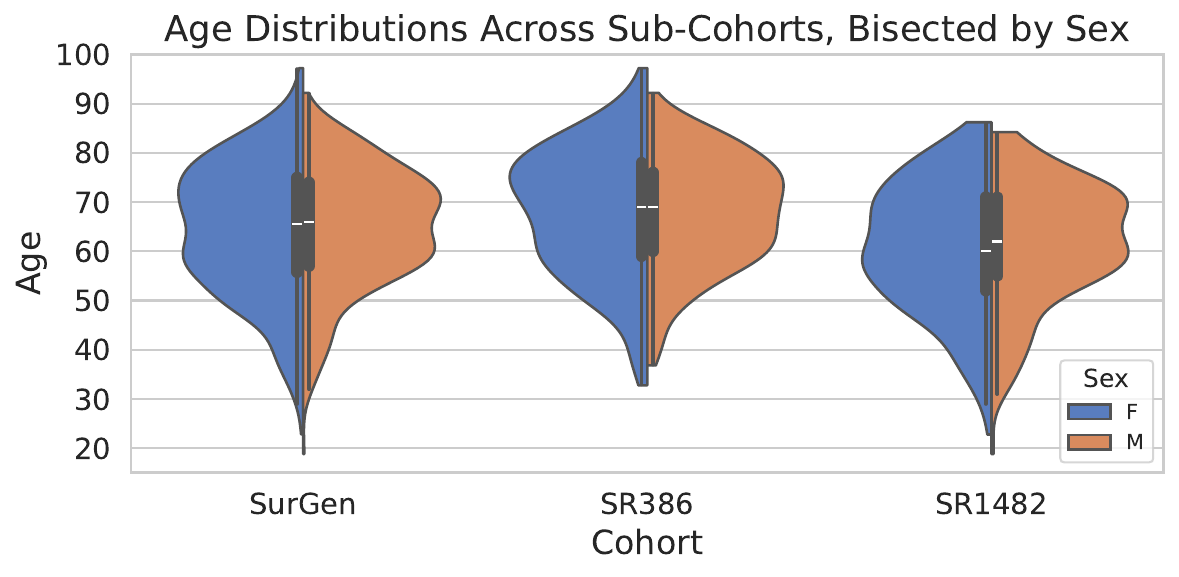}
\caption{Illustration of the age distributions (in years) of patients in each cohort, split by sex. The width of each violin represents the density of data points at different ages, highlighting the distributions within and across the cohorts.}\label{fig:violin-plot-demographic}
\end{figure}

The SurGen dataset comprises clinical information from 843 cases, with patients ranging from 19 to 97 years of age (mean age = 64.58, SD = \(\pm\)12.73), as illustrated in Figure \ref{fig:violin-plot-demographic}. The cohort comprises 46\% females and 54\% males.

\subsection{Patient Survival}

Survival data is available for the SR386 cohort, providing insights into patient outcomes over a five-year period following diagnosis. The dataset includes binary labels indicating whether a patient survived beyond the duration of the study, as well as the number of days until death for those who did not. For patients who outlived the study period or whose survival extends beyond the recorded date, their exact number of days until death is not captured, resulting in right-censoring.

\begin{figure}[bt!] 
\centering
\includegraphics[width=0.7\linewidth]{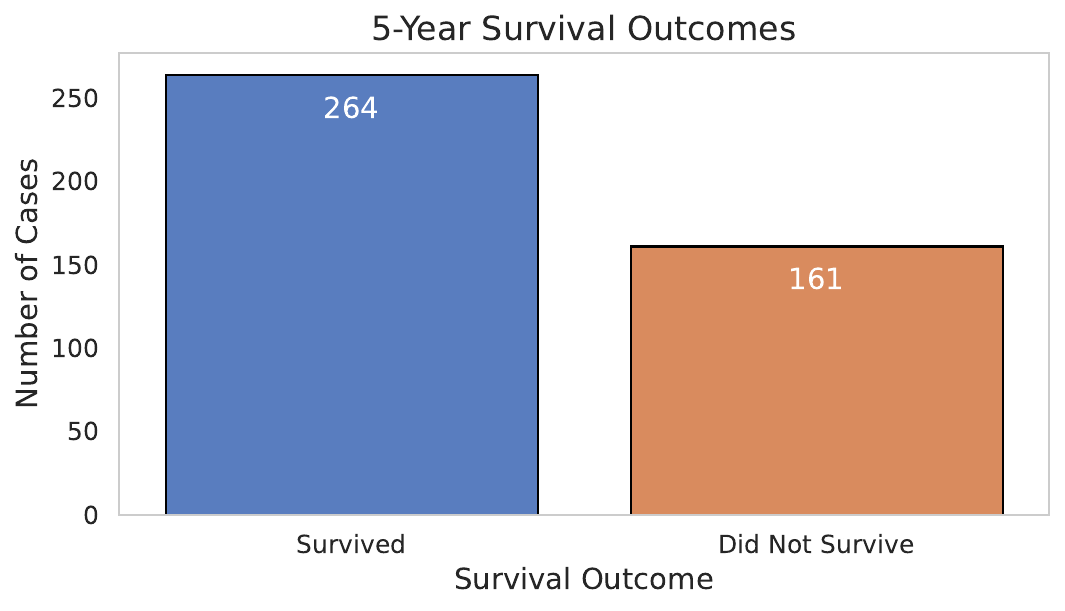}
\caption{Bar chart depicting the 5-year survival outcomes of the SR386 cohort. The chart shows the number of individuals who survived (n=264) versus those who did not survive (n=161) within the 5-year period following diagnosis. The data excludes instances where survival status was not recorded (NULL values).}\label{fig:bar-chart-surv}
\end{figure}

Within the SR386 cohort, 161 patients (38\%) died during the study period, while 264 patients (62\%) were alive at the end of the study period. This distribution provides a general overview of patient outcomes in the cohort. An overview of the binarised five-year survival outcomes is presented in Figure \ref{fig:bar-chart-surv}. CRC was the primary cause of death in 67 out of 161 deceased patients, accounting for 41.61\% of all deaths in the cohort.

\begin{figure}[bt!] 
\centering
\includegraphics[width=0.7\linewidth]{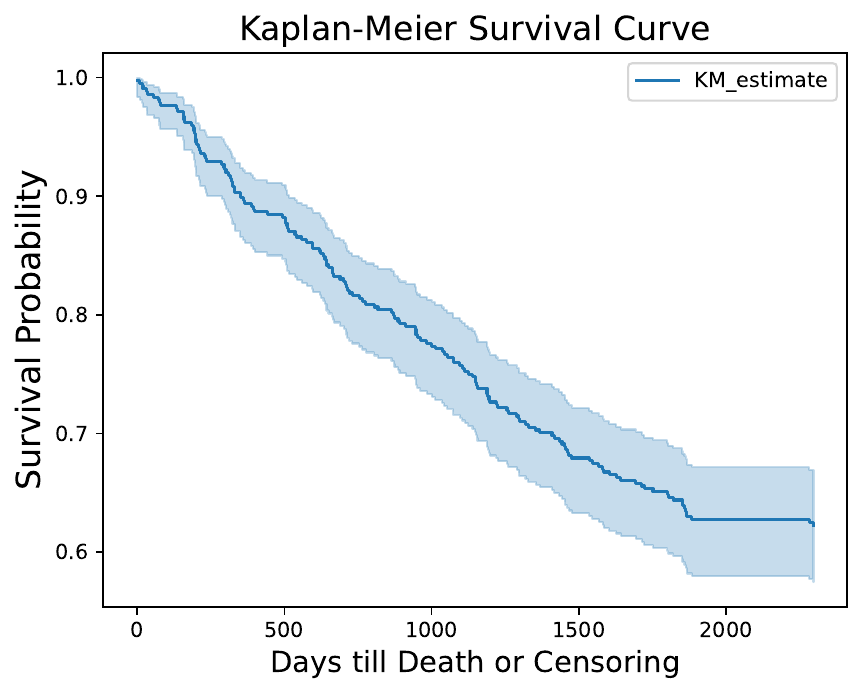}
\caption{Kaplan-Meier survival curve illustrating estimated survival probabilities over time. Censoring occurred for patients who survived beyond the 5-year study duration, as they were not followed further. The curve reflects the proportion of individuals surviving at each time point, with 95\% confidence intervals representing the uncertainty in these estimates.}\label{fig:kaplan-meirer-survival-curve}
\end{figure}

To visualise the survival probabilities over time, a Kaplan-Meier survival curve was constructed for the SR386 cohort, as shown in Figure \ref{fig:kaplan-meirer-survival-curve}. This curve illustrates the proportion of patients surviving at each time point during the study period. The gradual decline in the curve represents the decreasing number of patients alive as time progresses.

\begin{figure}[bt!] 
\centering
\includegraphics[width=0.7\linewidth]{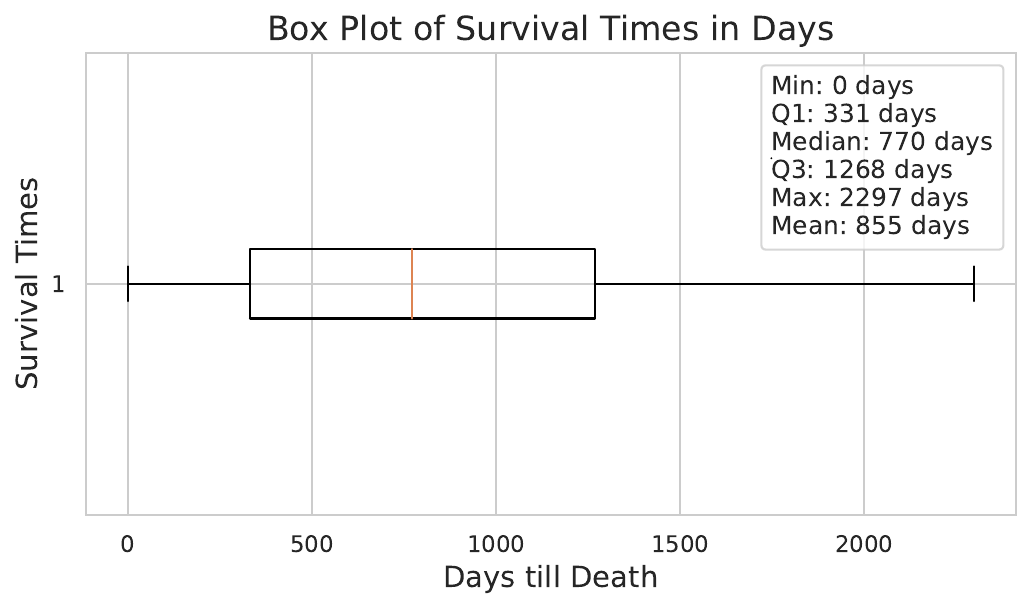}
\caption{Box plot showing the distribution of survival times (in days) for cases in the SR386 cohort with recorded days till death. The plot illustrates key summary statistics, including the mean, minimum, first quartile (Q1), median, third quartile (Q3), and maximum survival times.}\label{fig:box-plot-surv}
\end{figure}

For the patients who did not survive beyond the study period, we analysed the distribution of their survival times. Figure \ref{fig:box-plot-surv} presents a box plot summarising key statistics of these survival times in days. The plot shows the minimum, first quartile (Q1), median, third quartile (Q3), maximum, and mean survival times. Specifically, the median survival time was 770 days, indicating that half of the patients who died did so within this number of days post-diagnosis.

\begin{figure}[bt!] 
\centering
\includegraphics[width=0.7\linewidth]{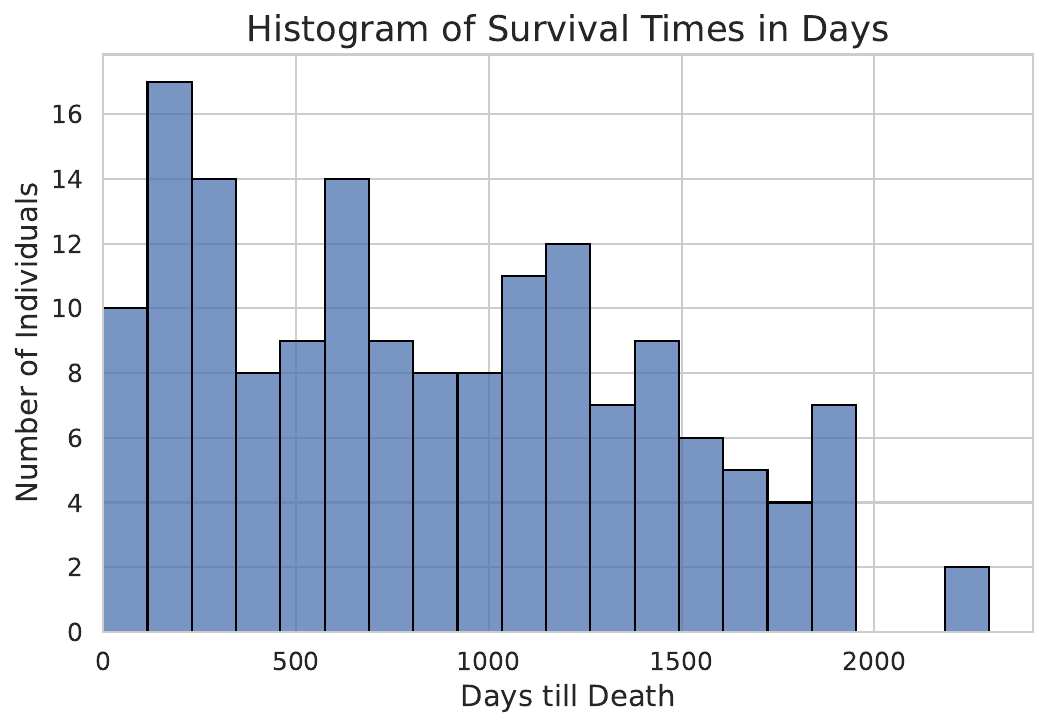}
\caption{Histogram showing the distribution of survival times (in days) for cases in the SR386 cohort with recorded days until death. The x-axis represents the number of days until death, and the y-axis indicates the number of individuals who died within each time interval.}\label{fig:histogram-surv}
\end{figure}

Additionally, Figure \ref{fig:histogram-surv} displays a histogram of the survival times for patients who died within the study period. The histogram shows how many patients died within specific time intervals, providing an overview of the distribution of survival times among these patients.

Due to quality control measures, missing information, or data inconsistencies, certain cases (i.e. 004, 208, 430) have been redacted or marked as `\textit{NULL}' with respect to survival. However, these cases remain in the dataset as they contain valuable genetic information that can be utilised for separate predictive tasks.

\subsection{Genetic Mutations}

\begin{figure}[bt!] 
\centering
\includegraphics[width=1\linewidth]{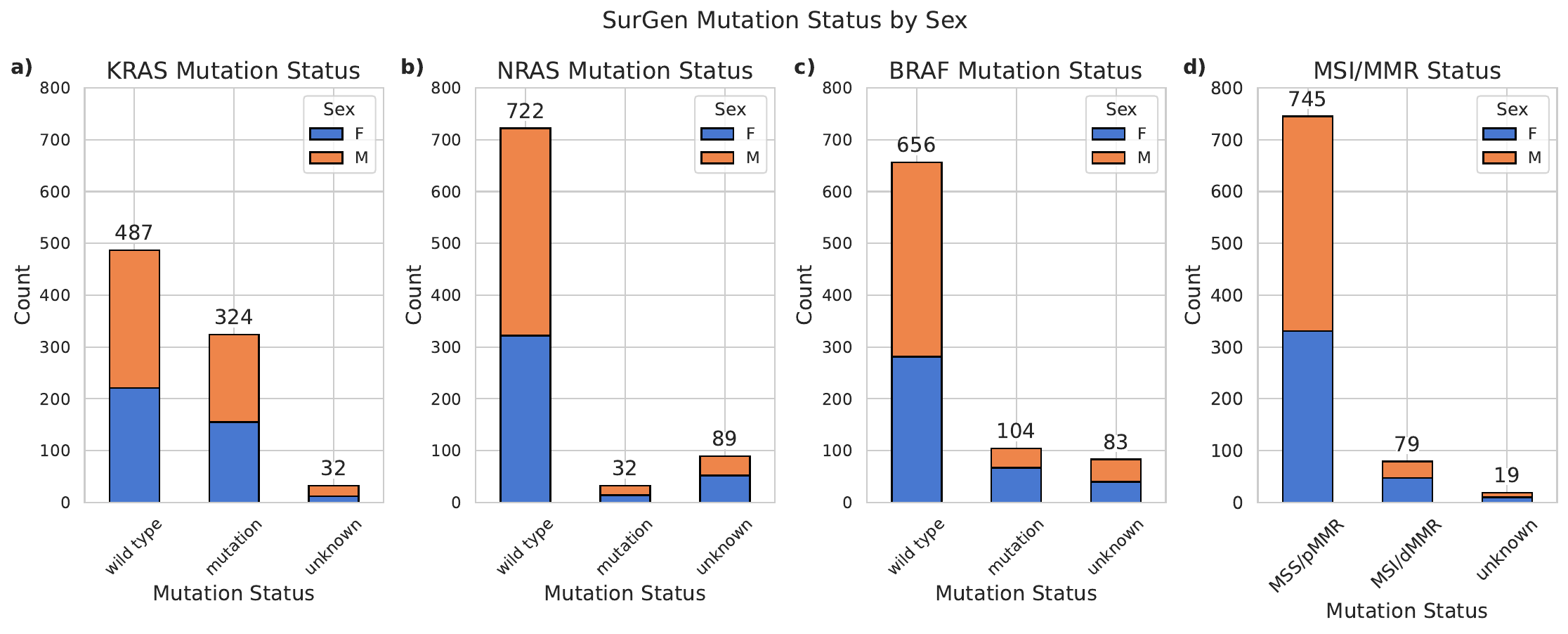}
\caption{Bar chart depicting mutation prevalence across the SurGen dataset, highlighting mutation status of patients across a) KRAS, b) NRAS, c) BRAF, and d) MSI/MMR.}\label{fig:mutation-plots}
\end{figure}

The SurGen dataset includes ground truth labels for key genetic mutations in the \textit{KRAS}, \textit{NRAS}, and \textit{BRAF} genes, as well as mismatch repair (MMR) status and/or microsatellite instability (MSI). Figure \ref{fig:mutation-plots} presents the distribution of these genetic mutations by sex. These genetic markers are crucial for understanding the molecular characteristics of tumours and their potential response to targeted therapies. Below, each mutation is discussed in detail.

\textbf{BRAF Mutation:} Present in 12.34\% of SurGen cases, aligning with frequencies reported in the literature, which range from 3.5\% to 13\% \citep{ogino2007evaluation, mirzapoor2023kras, guo2019clinicopathologic, de2010effects, sclafani2020analysis}. BRAF mutations are critical in the MAPK/ERK signalling pathway and are significant targets for therapeutic intervention \citep{burotto2014mapk, mccain2013mapk}.

\textbf{KRAS Mutation:} Present in 38.43\% of SurGen cases, consistent with the range reported in other studies, from 37\% to 46.4\% \citep{li2020braf, ogino2007evaluation, mirzapoor2023kras, guo2019clinicopathologic, de2010effects}. KRAS is a proto-oncogene involved in cell signalling pathways that regulate cell growth and death. Mutations in KRAS are often linked to resistance to specific therapies, highlighting the importance of their identification for effective treatment planning \citep{li2020braf}.

\textbf{NRAS Mutation:} Observed in 3.80\% of SurGen cases, this falls within the range of 2.6\% to 9\% reported across various studies \citep{mirzapoor2023kras, sclafani2020analysis, guo2019clinicopathologic, de2010effects}. Like KRAS, NRAS mutations can influence treatment options and prognosis, though NRAS mutations are less common.

\subsection{Mismatch Repair Deficiency and Microsatellite Instability}

Mismatch repair deficiency (dMMR) and microsatellite instability (MSI) are critical genetic features in many cancers, particularly colorectal cancer \citep{boland2010microsatellite}. dMMR occurs when the mismatch repair system, which normally corrects DNA replication errors, is compromised. This deficiency leads to an accumulation of mutations, particularly in regions of repetitive DNA known as microsatellites. When these microsatellites become unstable due to dMMR, the condition is termed microsatellite instability (MSI)\citep{vilar2010microsatellite, boland2010microsatellite}.

MSI is a key biomarker used to assess cancer prognosis and predict responses to certain therapies, such as immunotherapy. Tumours exhibiting high levels of MSI (MSI-high) are often associated with a better prognosis and may respond favourably to immune checkpoint inhibitors \citep{le2017mismatch, luchini2019esmo}. Identifying MMR and MSI status is essential for developing targeted treatment strategies and improving patient outcomes.

Importantly, dMMR and MSI are hallmark features of Lynch syndrome (LS), the most common hereditary colorectal cancer predisposition syndrome, accounting for approximately 3\% of all colorectal cancers \citep{lynch2009review, tiwari2016lynch}. LS, also known as hereditary non-polyposis colorectal cancer (HNPCC), is caused by germline mutations in the MMR genes (\textit{MLH1}, \textit{MSH2}, \textit{MSH6}, and \textit{PMS2}) \citep{hampel2005screening}, leading to a higher risk of developing colorectal cancer and other cancers at a younger age. Identifying patients with dMMR/MSI can therefore aid in diagnosing Lynch syndrome and facilitating genetic counselling \citep{lynch2008hereditary}.

In our study, the assessment of MMR status and MSI status differed between the SR386 and SR1482 cohorts.

\subsubsection{Assessment of MMR and MSI Status in Cohorts}

While the SR386 cohort reports MMR status assessed through immunohistochemistry (IHC) for key MMR proteins, the SR1482 cohort reports both MMR and MSI status.

\paragraph{SR386 Cohort}

In the SR386 cohort, MMR status was assessed exclusively using immunohistochemistry (IHC) for two key MMR proteins; MLH1 and PMS2. Cases were labelled according to the specific loss of expression observed. Primary antibodies against MLH1 and PMS2 were applied, and loss of nuclear staining in tumour cells for any of these MMR proteins was recorded.

\paragraph{SR1482 Cohort}

In the SR1482 cohort, MSI status was determined using either immunohistochemistry (IHC) for MMR proteins (MLH1, MSH2, MSH6, PMS2) or PCR-based fragment analysis. For the PCR-based approach, the Promega Oncomate\textsuperscript{\texttrademark} kit was utilised according to the manufacturer's recommended protocol. Cases were classified as MSI/dMMR if they showed evidence of microsatellite instability through PCR analysis or a loss of protein expression by IHC.

\subsubsection{Mismatch Repair}

Mismatch repair (MMR) status is available for most cases, with a distinction between microsatellite stable (MSS/pMMR) and microsatellite unstable (MSI/dMMR) tumours within the SR1482 dataset. This information is crucial for identifying patients who might benefit from immunotherapy \citep{le2017mismatch, vilar2010microsatellite}.

\subsubsection{Microsatellites}

Microsatellite instability (MSI) is a condition of genetic hypermutability that results from impaired DNA mismatch repair (MMR). Identifying MSI is important as it has implications for the prognosis and treatment of cancer.

\subsection{Staging}

Tumour staging is a critical aspect of cancer diagnosis and treatment planning, providing a framework for assessing the extent of cancer spread within the body. Staging systems help in predicting patient prognosis, guiding treatment decisions, and enabling comparisons across clinical studies and populations \citep{amin2017eighth}. Two widely used staging systems in colorectal cancer are the Dukes' staging system\citep{dukes1932classification} and the TNM (Tumour, Node, Metastasis) staging system\citep{international1974tnm}, each offering distinct advantages and serving different clinical needs.

The Dukes' staging system is one of the earliest methods used to classify the extent of colorectal cancer. It is relatively simple and easy to apply, making it useful for broad clinical assessments. However, although it includes stages for lymph node involvement (Stage C) and distant metastasis (Stage D), its simplicity limits its ability to provide more detailed, granular information on tumour characteristics \citep{haq2009dukes}.

The TNM staging system, in contrast, is more detailed and widely applicable across various cancer types. It provides a comprehensive classification based on the size and extent of the primary tumour (T), the involvement of regional lymph nodes (N), and the presence of distant metastasis (M). This system is advantageous for its specificity and adaptability to different cancers, though it can be more complex to use compared to the Dukes' system.

These staging systems are integral to clinical guidelines, informing treatment strategies such as surgical intervention, chemotherapy, and targeted therapies based on the stage of cancer.

The SurGen dataset includes tumour staging information using both the Dukes' and TNM staging systems, which are essential for correlating clinical outcomes with tumour progression. Understanding the distribution of these stages across the cohort can offer valuable insights into the disease dynamics within the study population.

\subsubsection{Tumour Staging with Dukes'}

The Dukes' staging system classifies colorectal cancer into four stages (A, B, C, and D), based on the extent of tumour invasion and the presence of lymph node involvement or distant metastasis \citep{dukes1932classification}. Stage A represents the earliest form of cancer, confined to the mucosa, while Stage D indicates advanced disease with distant metastasis. This system, though less detailed than TNM, provides a quick and accessible way to gauge tumour progression and patient prognosis.

\subsubsection{TNM Staging}

The TNM staging system is a more granular approach that classifies cancer based on three key components: the size and extent of the primary tumour (T), the involvement of regional lymph nodes (N), and the presence of distant metastasis (M) \citep{sobin2011tnm}. Each of these components is assigned a score, and the combination of these scores determines the overall stage of the cancer, ranging from Stage 0 (in situ, non-invasive cancer) to Stage IV (advanced cancer with distant metastasis).

\noindent A comprehensive summary of the SurGen including survival data, genetic mutations, and image properties for both the SR386 and SR1482 sub-cohorts is provided in Table \ref{tab:surgen-clinical-overview}.

\begin{center}
\begin{table*}[]
\caption{Overview of the SurGen dataset with respective technical, clinical, and mutational characteristic breakdown of the sub-sets SR386 and SR1482. Note: MSI/MMR ground truth was determined using Immunohistochemistry (IHC) or Polymerase Chain Reaction (PCR).}
\label{tab:surgen-clinical-overview}
\begin{tabular}{llll}
                                        & \textbf{SurGen Dataset} & SR386            & SR1482                  \\ \hline
Origin                                  & Scotland                & Scotland         & Scotland                \\ \hline
Number of cases                         & 843                     & 427              & 416                     \\
Number of WSIs                          & 1020                    & 427              & 593                     \\
WSI file format                         & .CZI                    & .CZI             & .CZI                    \\
Magnification                           & 40X                     & 40X              & 40X                     \\
Microns per pixel (pixel width)         & 0.1112$\mu m$           & 0.1112$\mu m$    & 0.1112$\mu m$           \\ \hline
Mean age (std. dev.)                    & 64.58 (±12.73)          & 67.89 (±12.00)   & 61.20 (±12.59)          \\ \hline
Female, n (\%)                          & 388 (46.03\%)           & 197 (46.14\%)    & 191 (45.91\%)           \\
Male, n (\%)                            & 455 (53.97\%)           & 230 (53.86\%)    & 225 (54.09\%)           \\ \hline
MSI/MMR ground truth                    & PCR/IHC                     & IHC & PCR/IHC \\
MSI/dMMR, n (\%)                        & 79 (9.37\%)             & 32 (7.49\%)      & 47 (11.30\%)            \\
MSS/pMMR, n (\%)                        & 745 (88.37\%)           & 395 (92.51\%)    & 350 (84.13\%)           \\
MSI/MMR status unknown, n (\%)          & 19 (2.25\%)             & 0 (0\%)          & 19 (4.57\%)             \\ \hline
Five year survival (true), n (\%)       & 264 (31.32\%)           & 264 (61.83\%)    & 0 (0\%)                 \\
Five year survival (false), n (\%)      & 162 (19.22\%)           & 162 (37.94\%)    & 0 (0\%)                 \\
Five year survival (unreported), n (\%) & 417 (49.47\%)           & 1 (0.23\%)       & 416 (100\%)             \\ \hline
BRAF mutation, n (\%)                   & 104 (12.34\%)           & 47 (11.00\%)     & 57 (13.70\%)            \\
BRAF wild type, n (\%)                  & 656 (77.82\%)           & 379 (88.76\%)    & 277 (66.59\%)           \\
BRAF status unknown, n (\%)             & 83 (9.85\%)             & 1 (0.23\%)       & 82 (19.71\%)            \\ \hline
KRAS mutation, n (\%)                   & 324 (38.43\%)& 147 (34.43\%)& 177 (42.55\%)\\
KRAS wild type, n (\%)                  & 487 (57.77\%)& 266 (62.30\%)& 221 (53.12\%)\\
KRAS status unknown, n (\%)             & 32 (3.80\%)& 14 (3.26)& 18 (4.33\%)\\ \hline
NRAS mutation, n (\%)                   & 32 (3.80\%)             & 16 (3.75\%)      & 16 (3.85\%)             \\
NRAS wild type, n (\%)                  & 722 (85.65\%)           & 399 (93.44\%)    & 323 (77.64\%)           \\
NRAS status unknown, n (\%)             & 89 (10.56\%)            & 12 (2.81\%)      & 77 (18.51\%)            \\ \hline
\end{tabular}
\end{table*}
\end{center}

\subsection{Data collection}

\subsubsection{Tissue Sample Preparation}

Samples underwent formalin-fixed-paraffin-embedding (FFPE) processing. This involved fixing tissue specimen in formalin to preserve cellular structures and proteins, followed by embedding the samples in paraffin wax.

Once FFPE samples were prepared, they were processed using a microtome set to section at 5$\mu m$ before being laid onto a glass slide. These slides were then subjected to routine haematoxylin and eosin (H\&E) staining prior to their digitisation.

Slides were first immersed in haematoxylin, which stains the cell nuclei blue-purple. Following a rinse, slides were stained with eosin, which stains the cytoplasm and extracellular matrix pink. After staining, the slides underwent a dehydration process involving graded alcohols and xylene. Coverslips were subsequently applied with a mounting medium to preserve the stained sections.

\subsubsection{Tissue Sample Digitisation}

\begin{figure}[bt!]
\centering
\includegraphics[width=0.7\linewidth]{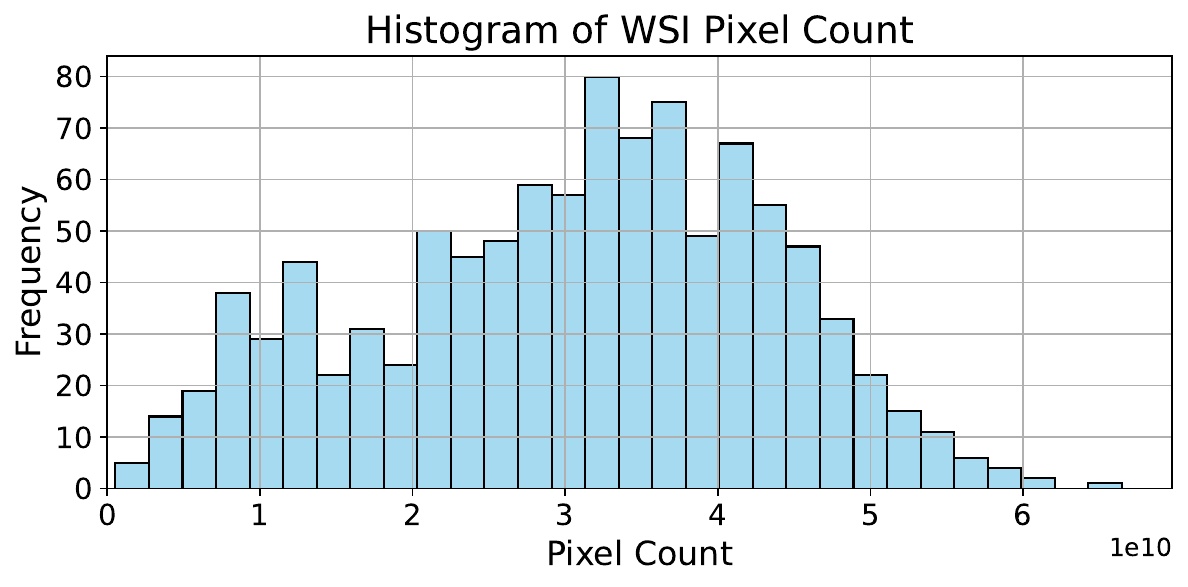}
\caption{Histogram illustrating the scale of SurGen whole slide images with respect to the number of pixels per image. The x-axis represents the total number of pixels in each image (in tens of billions, $1\times10^{10}$), while the y-axis indicates the frequency of occurrence of images within each bin. The distribution shows the variability in the sizes of whole slide images across the dataset.}\label{fig:pixel-count-histogram}
\end{figure}

Prepared slides were digitised on-site using a \textit{ZEISS Axio Scan.Z1 Microscopy Slide Scanner} at 40$\times$ magnification, equipped with a Plan-Apochromat 40x/0.95 Korr M27 objective lens. This combination produces a pixel size of $0.1112\mu m$. The scans were performed using ZEN 2.6 (blue edition) software, capturing brightfield images with controlled transmitted light illumination. Digitised images were saved in 24-bit BGR format (BGR24) with a pixel size of $0.1112\mu m$. A multi-resolution pyramidal image structure was generated, with each subsequent layer downsampled by a factor of 2 relative to the previous layer, using Gaussian filtering to maintain image quality. Figure \ref{fig:pixel-count-histogram} illustrates the WSI pixel counts across the SurGen dataset. 

While this imaging setup yields an ultra-high resolution of $0.1112\,\mu m$ per pixel, it is important to note that pixel size is not standardised across all digital pathology platforms. Scanners from other vendors (e.g., Philips, Hamamatsu, Leica) may report the same nominal magnification (e.g., 40$\times$), yet produce images with coarser resolutions, typically around $0.25\,\mu m$ per pixel, due to differences in objective lens, camera sensors, and internal optics. As such, magnification alone is an imprecise descriptor of image resolution. Reporting the true microns-per-pixel (MPP) value provides a more objective and reproducible measure of resolution, enabling proper normalisation across datasets acquired using differing hardware configurations.

For both cohorts, each whole slide image was digitised from the same formalin-fixed paraffin-embedded (FFPE) tissue block used for biomarker assessment, ensuring correspondence between the histological and molecular data.

\subsubsection{DNA Sequencing}

Next Generation Sequencing (NGS) was performed to determine the mutation status of KRAS, NRAS, and BRAF using the Ion Torrent\textsuperscript{\texttrademark} Cancer Hotspot Panel v2 (Thermo Fisher Scientific), following the manufacturer's protocol.

\subsection{Data Curation and Quality Control}

To ensure the quality and reliability of the SurGen dataset, we implemented several data curation and quality control measures.

\subsubsection{Slide Quality Assessment}

All WSIs were reviewed by specialised laboratory personnel trained in the preparation of tissue samples for microscopic examination. Each slide was assessed for staining quality, focus, and absence of artifacts. Slides that did not meet acceptable standards were re-scanned or re-prepared to improve image quality.

\subsubsection{Data Alignment and Consistency}

To maintain data integrity and maximise the utility of the dataset, we carefully matched each WSI with its corresponding clinical and genetic data. WSIs without any matching clinical data were excluded from the dataset, as clinical context is essential for meaningful analyses. However, clinical data entries were retained even if some fields were incomplete, provided they had a corresponding WSI. This approach ensured that all included WSIs had associated clinical information, enhancing the dataset's applicability while acknowledging that some clinical records might have missing data points.

\subsubsection{Anonymisation and Ethical Considerations}

Patient confidentiality was prioritised throughout the curation of the SurGen dataset. In line with contemporary data ethics in computational pathology \citep{abels2019computational, holub2023privacy}, we implemented deidentification protocols to ensure privacy while maximising data utility. Recognising that medical images potentially carry the risk of re-identification when combined with external data sources, our anonymisation strategy involved the removal or redaction of potentially identifiable information, including dates of diagnosis, date of death, and treatment details.

\subsection{Data Use}

Researchers can interact with the WSIs using tools such as OpenSlide \citep{goode2013openslide}, pylibCZIrw \citep{zeiss_pylibczirw}, and Bioformats \citep{linkert2010metadata}. The images are saved in a hierarchical pyramidal format, facilitating efficient viewing and processing at multiple resolutions. Software such as QuPath \citep{bankhead2017qupath}, Fiji \citep{schindelin2012fiji}, ImageJ \citep{schneider2012nih}, and others can be used to visualise and analyse these images.

\begin{figure}[bt!]
\centering
\includegraphics[width=0.5\linewidth]{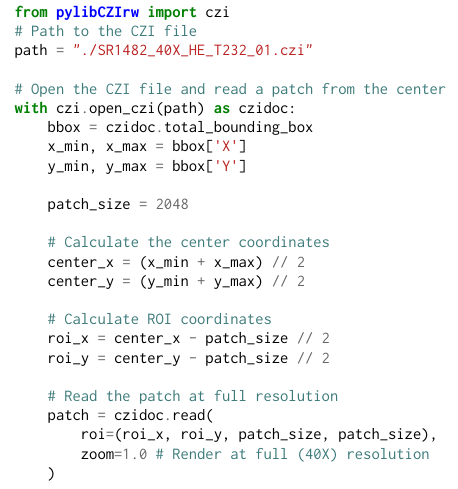}
\caption{Python code demonstrating how to extract a tile from the centre of a WSI using in \texttt{Python 3.8.13} and \texttt{pylibCZIrw v4.1.3}. This example illustrates how to interact with high-resolution pathology images in CZI format. This method can be easily expanded to tessellate over an entire whole slide image for the purpose of patch-level feature extraction.}\label{fig:tile-extraction-example}
\end{figure}

\begin{figure}[b!] 
\centering
\includegraphics[width=0.4\linewidth]{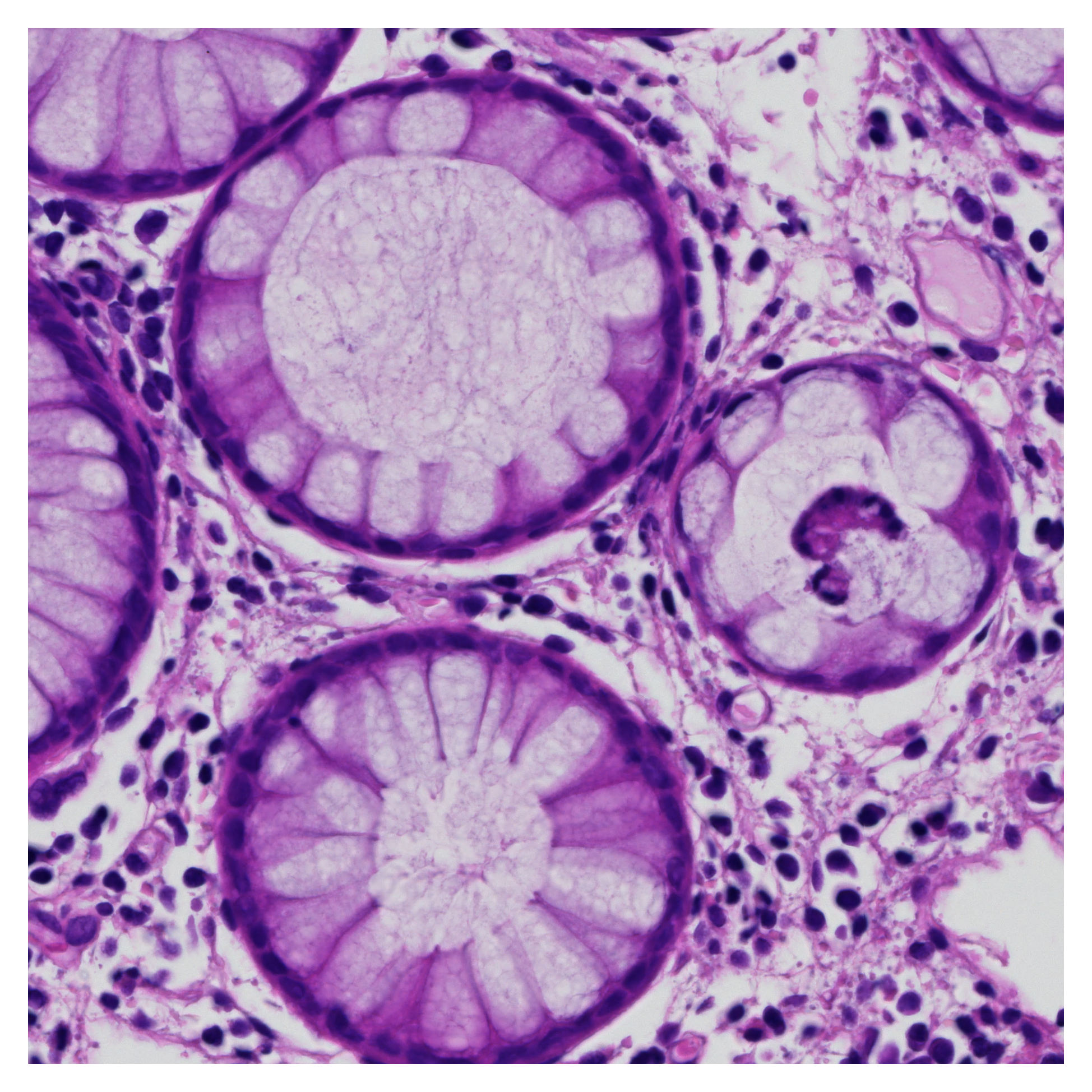}
\caption{Example 2048 × 2048 pixel tile extracted from the centre of a whole slide image (WSI) using \texttt{pylibCZIrw}. This patch, from case SR1482\_T232, illustrates the fine detail captured at 40X (0.1112$\mu m$ per pixel) resolution. Extraction of patches, as demonstrated here, is an essential step in SOTA preprocessing pipelines.}\label{fig:tile-export-image}
\end{figure}

To illustrate SurGen's practical utility, we provide a simple Python example for extracting a region of interest from a whole slide image. A Python script was implemented using \texttt{pylibCZIrw} (See Figure~\ref{fig:tile-extraction-example}). The script illustrates the process of identifying the centre of the WSI and extracting a 2048 × 2048 pixel region of interest (ROI) at full resolution. The extracted tile (Figure \ref{fig:tile-export-image}) provides a high-resolution view from the WSI, showcasing the potential for downstream analyses or tasks, such as patch-level feature extraction or visualisation.

\subsection{Data re-use Potential}

The SurGen dataset offers extensive opportunities for researchers in computational pathology and oncology. Its comprehensive collection of WSIs, coupled with genetic and other clinical annotations, makes it a valuable resource for various applications.

Firstly, the dataset can be utilised to train machine learning models for predicting mismatch repair (MMR) status and microsatellite instability (MSI). Given that existing publicly available datasets focusing on MSI/MMR prediction are limited, SurGen fills a crucial gap. Researchers can leverage this dataset to develop and validate models that may enhance diagnostic accuracy and inform treatment strategies, particularly in colorectal cancer where MSI status is a key prognostic and therapeutic marker.

Secondly, SurGen provides a rich resource for training models aimed at genomic mutation prediction, specifically for mutations in the KRAS, NRAS, and BRAF genes. Expanding the quantity of publicly available datasets with such detailed genetic information is immensely valuable, as it enables the development of models that can predict genetic mutations from histopathological images. This can potentially streamline the diagnostic process by reducing the need for costly and time-consuming genetic testing.

Furthermore, the high-quality WSIs in the SurGen dataset make it suitable for training foundation models in digital pathology. Existing works have demonstrated that the performance of these models improves with the availability of larger and more diverse datasets \citep{chen2024towards, oquab2023dinov2, oliveira2021cad}. By contributing to the training of such models, SurGen can aid in advancing the field of computational pathology, facilitating the development of algorithms that are more robust and generalisable.

The dataset's versatility allows it to be used in multiple ways:

\begin{itemize}
    \item Researchers may choose to utilise the SR386 or SR1482 subsets independently, depending on their specific research questions. For instance, studies focusing on primary tumour characteristics and survival can benefit from the SR386 cohort's valuable genetic and survival data.

    \item Alternatively, the entire SurGen dataset can be employed collectively as a larger cohort for tasks such as staging or genetic slide-level classification, benefiting from the increased sample size and additional diversity from metastatic tumour sites.

    \item SurGen also holds significant potential as an external validation set for existing studies and algorithms. External validation is essential for assessing the generalisability of predictive models, and the dataset's comprehensive annotations make it particularly suitable for this purpose \citep{cui2021artificial}.
\end{itemize}

To support systematic benchmarking and methodological comparisons, we provide example stratified data-splits for the SR386 subset (see Table \ref{tab:SR386-data-distribution}), as well as for the SR1482 subset and the combined SurGen dataset. Although detailed stratifications are only presented here for SR386, equivalent splits for the full SurGen dataset and the SR1482 subset are available in the accompanying GitHub repository. Each split is stratified to ensure balanced distributions of key variables such as genetic mutations, MMR/MSI status, and survival metrics. These data partitions establish a standardised, transparent framework for evaluating model performance and reproducibility when utilising the SurGen dataset.

An example of the dataset's utility is demonstrated in a study that explored the feasibility of digital pathology foundation models on the SR386 cohort. Using the UNI model \citep{chen2024towards}, which was benchmarked against various other pathology-pretrained foundation models and an ImageNet-pretrained ResNet-50 \citep{he2016deep}, this work achieved a test AUROC of 0.7136 for slide-level classification of MMR status \citep{myles2024leveraging}. This underscores the dataset's potential in facilitating advanced machine learning applications.

\begin{table*}[bt!]
\caption{Breakdown of SR386 SurGen Colorectal Cohort data distribution for train, validate, and test sets. This stratification may act as an effective starting point for future analysis. Each patient has precisely one associated whole slide image. This breakdown was stratified by age, sex, MSI/MMR, RAS (KRAS or NRAS), and BRAF mutation.}
\label{tab:SR386-data-distribution}
\begin{tabular}{@{}l|llll@{}}
\toprule
Category              & Total (SR386)& Train        & Validate    & Test        \\ \midrule
Origin                & Scotland     & Scotland     & Scotland    & Scotland    \\
WSI file format       & CZI          & CZI          & CZI         & CZI         \\
Magnification         & ×40          & ×40          & ×40         & ×40         \\
Microns per pixel (pixel width)    & 0.1112$\mu m$  & 0.1112$\mu m$  & 0.1112$\mu m$  & 0.1112$\mu m$  \\ \midrule
Number of patients    & 423 (100\%)  & 255 (60\%)   & 84 (20\%)   & 84 (20\%)   \\ 
Mean age at diagnosis (std. dev.)  & 67.89 (±11.97) & 67.98 (±12.12) & 67.71 (±11.40) & 67.80 (±12.20) \\ \midrule
Male, n (\%)          & 228 (54\%)   & 138 (54.1\%) & 46 (54.7\%) & 44 (52.3\%) \\
Female, n (\%)        & 195 (46.0\%) & 117 (45.8\%) & 38 (45.2\%) & 40 (47.6\%) \\ \midrule
MSS/pMMR, n (\%)      & 391 (92\%)   & 235 (92\%)   & 78 (93\%)   & 78 (93\%)   \\
MSI/dMMR, n (\%)      & 32 (8\%)     & 20 (8\%)     & 6 (7\%)     & 6 (7\%)     \\ \midrule
Five year survival (true), n (\%)  & 159  (38\%)    & 100 ( 39\%)    & 30  (36\%)     & 29  (35\%)     \\
Five year survival (false), n (\%) & 264  (62\%)    & 155  (61\%)    & 54  (64\%)     & 55  (65\%)     \\ \midrule
RAS mutation, n (\%)  & 158 (37\%)   & 97 (38\%)    & 31 (37\%)   & 30 (36\%)   \\
RAS wild type, n (\%) & 265 (63\%)   & 158 (62\%)   & 53 (63\%)   & 54 (64\%)   \\ \midrule
BRAF mutation, n (\%)              & 47 (11.1\%)    & 29 (11.4\%)    & 9 (10.7\%)     & 9 (10.7\%)     \\
BRAF wild type, n (\%)             & 375 (88.6\%)   & 225 (88.2\%)   & 75 (89.2\%)    & 75 (89.2\%)    \\
BRAF fail, n (\%)     & 1 ( 0.2\%)   & 1  (0.4\%)   & 0 (0\%)     & 0 (0\%)     \\ \bottomrule
\end{tabular}%
\end{table*}

\section{Analyses}

To further demonstrate the utility of the SurGen dataset, we conducted an experiment combining the SR386 and SR1482 cohorts to predict MMR status using a machine learning model. We utilised the existing training, validation, and test splits from each cohort and merged them to form unified training, validation, and test sets. This approach ensured that the combined SurGen dataset adhered to the 60:20:20 ratio for training, validation, and testing, respectively, while maintaining a balanced representation of mutation statuses across each split. By leveraging the predefined splits from both cohorts, we eliminated the need to generate a separate third split. The splits used in this experiment are provided in CSV format to ensure reproducibility.

\subsection{Feature Extraction} 

A range of pre-trained foundation models have been developed for histopathological image analysis, each leveraging diverse self-supervised learning techniques and trained on extensive collections of WSIs. These models have demonstrated considerable success in capturing nuanced histopathological features \citep{wang2022transformer, azizi2023robust, chen2022scaling, kang2023benchmarking, filiot2023scaling, lu2024visual, chen2024towards, vorontsov2024foundation, campanella2023computational, lai2023domain, hua2024pathoduet, dippel2024rudolfv, aben2024towards, juyal2024pluto, yang2024foundation, xu2024whole, nechaev2024hibou, hoptimus0, xu2024multimodal, zimmermann2024virchow, filiot2024phikon, wang2024pathology, ding2024multimodal}.

For this study, we employed the UNI foundation model \citep{chen2024towards} for feature extraction from WSIs. UNI was selected due to its robust performance in representing histopathological features relevant to microsatellite instability (MMR status) within the SR386 cohort \citep{myles2024leveraging}. The model is a self-supervised vision encoder trained on over 100,000 H\&E-stained WSIs across a wide variety of tumour sites, thereby providing a comprehensive representation of tissue morphology.

\begin{figure}[bt!] 
\centering
\includegraphics[width=0.7\linewidth]{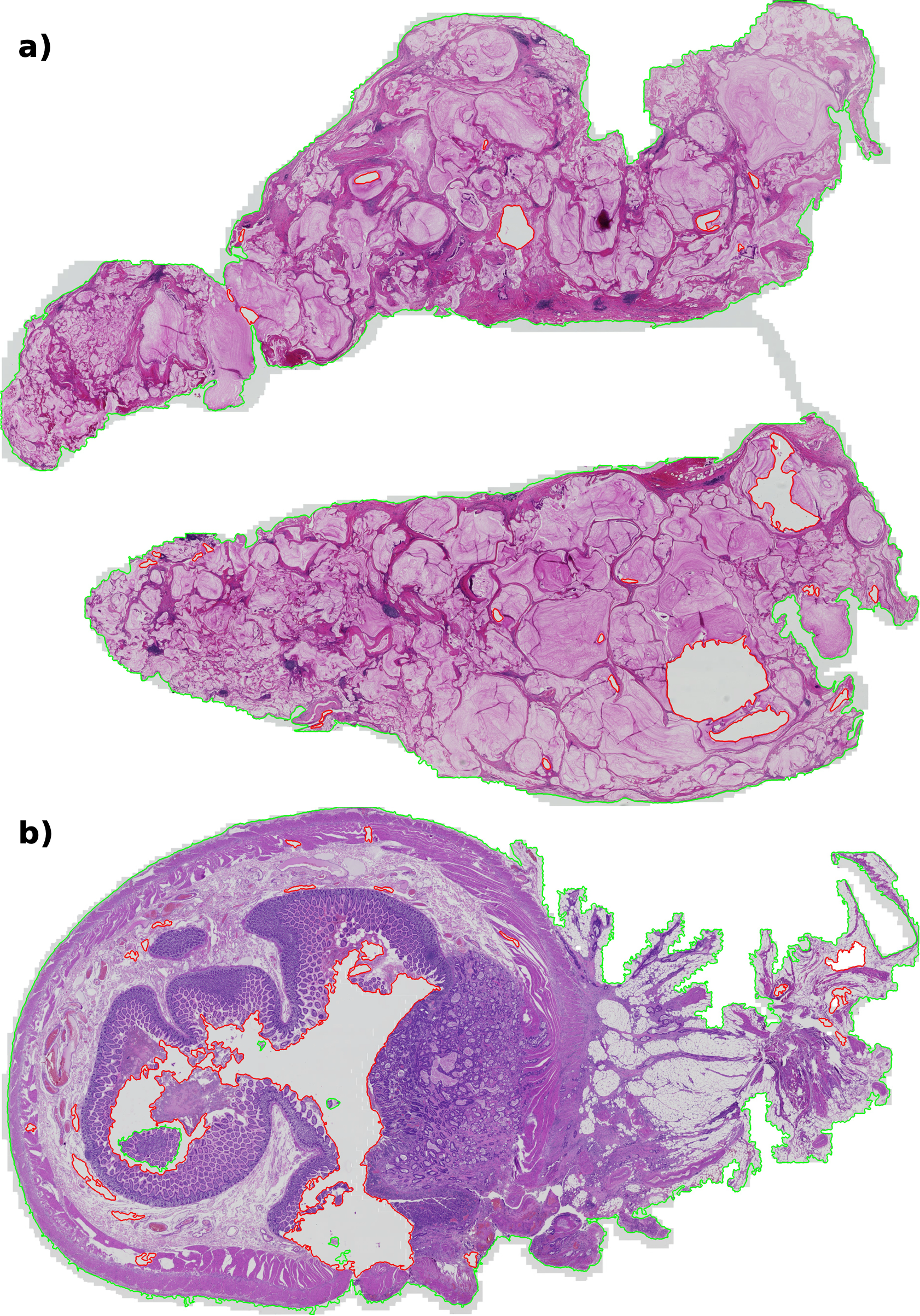}
\caption{Background subtraction from a) case SR148\_T230, peritoneal biopsy and b) case SR148\_T412, small bowel resection. Tissue area is outlined in green with holes and background delineated in red.}\label{fig:segmented-wsis}
\end{figure}

Feature extraction was performed on non-overlapping 224x224 tissue patches at a scale of 1.0 microns per pixel (MPP), yielding a 1024-dimensional embedding for each patch. Background subtraction was applied as illustrated in Figure \ref{fig:segmented-wsis}. The entire process of patch extraction and feature embedding required 110.55 hours, utilising a single NVIDIA V100 32GB GPU. For convenience and reproducibility, these embeddings are made available online.

\subsection{Model Training and Evaluation}

\begin{figure}[bt!] 
\centering
\includegraphics[width=0.7\linewidth]{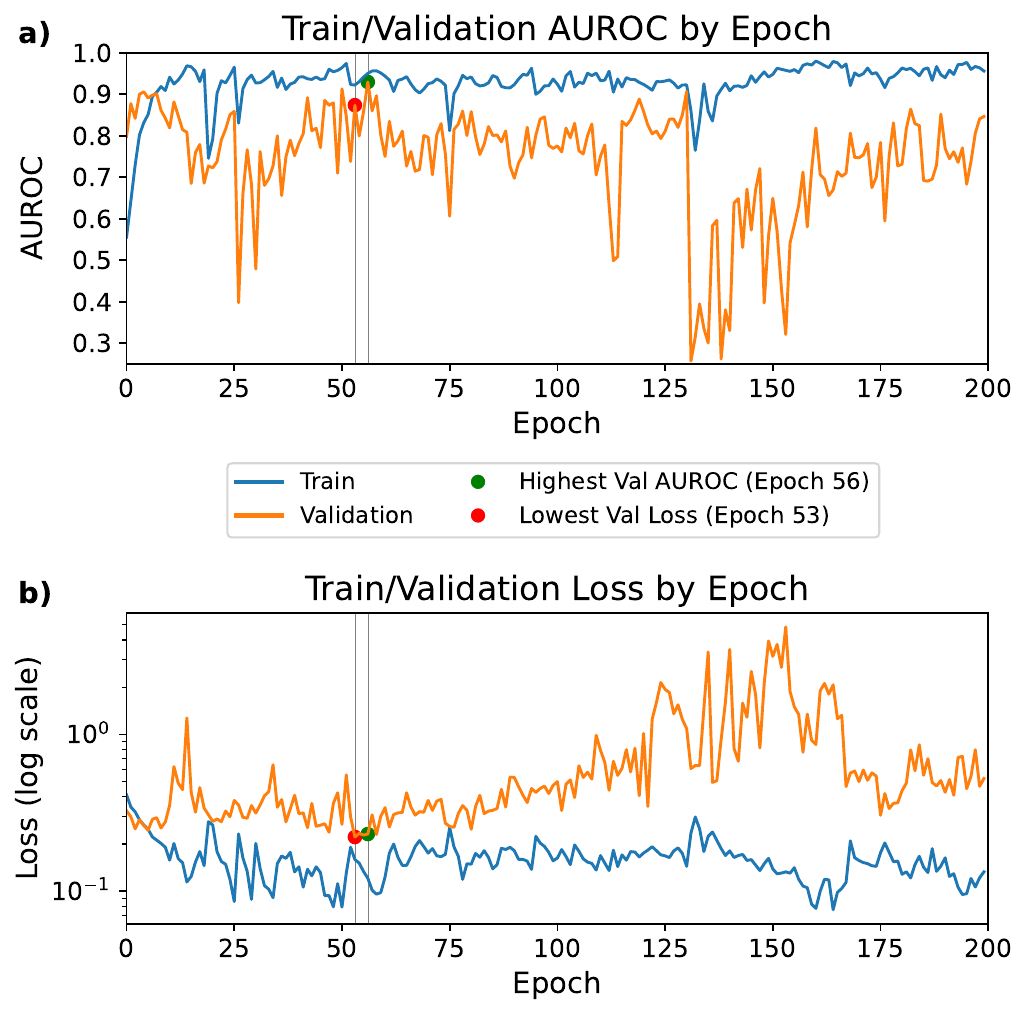}
\caption{Train/Validation AUROC and Loss by Epoch: a) illustrates the train and validation AUROC progression over 200 epochs, with markers indicating the highest validation AUROC and the epoch with the lowest validation loss. b) shows the train and validation loss on a log scale, highlighting the convergence and divergence trends, with markers indicating key performance metrics such as the lowest validation loss and the epoch with the highest AUROC.}\label{fig:train-val-plot}
\end{figure}

\begin{figure}[bt!] 
\centering
\includegraphics[width=0.5\linewidth]{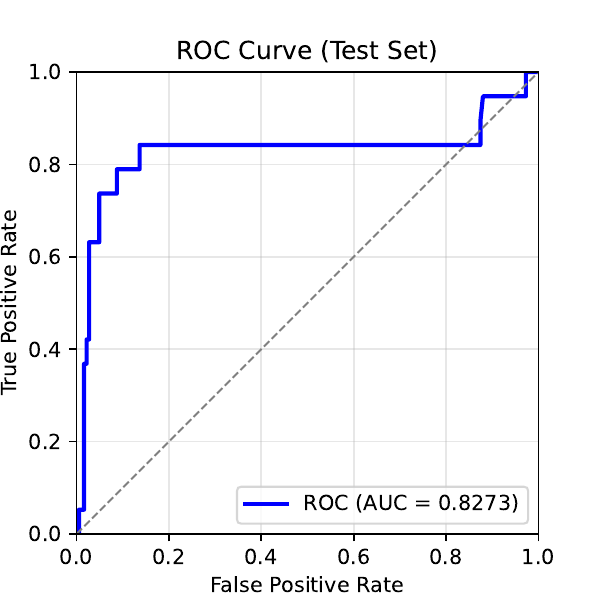}
\caption{Receiver Operating Characteristic (ROC) curve for the model, showing an AUROC of 0.8273. The curve plots the true positive rate (sensitivity) against the false positive rate (1 - specificity) across various classification thresholds, with an AUROC of 1 representing perfect classification and 0.5 indicating random chance.}\label{fig:test-roc-curve}
\end{figure}

A Transformer \citep{vaswani2017attention} based classifier was trained using the extracted UNI patch embeddings. Details of the model parameters are provided in Table \ref{tab:model-parameters}. Performance was evaluated primarily using the Area Under the Receiver Operating Characteristic curve (AUROC) metric. Training was conducted on a single NVIDIA V100 32GB GPU, completing in 3 hours, 2 minutes, and 36 seconds. The progression of the training and validation AUROC, as well as the loss over 200 epochs, is shown in Figure \ref{fig:train-val-plot}. This figure highlights key performance metrics, including the highest validation AUROC and the lowest validation loss. Preliminary results indicate a validation AUROC of 0.9297 and a test AUROC of 0.8273 (see Figure \ref{fig:test-roc-curve} for test AUROC curve). These results demonstrate the model's potential for accurately predicting MMR status from WSIs. Future work could focus on fine-tuning hyperparameters and exploring the integration of state-of-the-art (SOTA) pretrained feature extractors to further improve model performance.

\subsubsection{Model Architecture}

The model consists of a feature embedding layer, a transformer encoder, an aggregation layer, and a classification head. The feature extractor used was the UNI model, which produced 1024-dimensional feature vectors for each patch. These were mapped to a 512-dimensional latent space via a fully connected layer and ReLU activation. The transformer encoder consisted of 2 layers, each with 2 attention heads, and a feedforward dimension of 2048. After passing through the transformer encoder, the patch features were mean-pooled to obtain a slide-level feature representation. A final fully connected layer then mapped the pooled feature vector to the number of classes (for multi-class tasks) or to a single output (for binary classification). The full architecture configuration is detailed in table \ref{tab:model-parameters}.

\subsubsection{Training Configuration}

The model was trained using patch embeddings extracted from WSIs at 1.0$\mu/pixel$ per pixel, with patch sizes of 224x224. As the number of patches per WSI varied based on the specimen size, we processed all patches in a single forward pass. The training was conducted on a single NVIDIA V100 32GB GPU, with a batch size of 1 and a learning rate of $1 \times 10^{-4}$. The Adam optimiser was used, and binary cross-entropy with logits loss (\texttt{BCEWithLogitsLoss}) was applied for binary classification tasks. No class balancing was performed. The model was trained for 200 epochs, and automatic mixed precision (AMP) was enabled to optimise GPU usage. Table \ref{tab:model-parameters} provides a summary of the key parameters used in the training process.

\begin{table}[htbp]
\centering
\caption{Summary of model parameters used for MMR/MSI classification.}
\label{tab:model-parameters}
\begin{tabular}{ll} 
\toprule
\textbf{Parameter}                   & \textbf{Value}                         \\ 
\midrule
\textbf{Task}                        & MMR/MSI Detection                      \\ 
\textbf{Cohort}                      & SurGen                                 \\ 
\textbf{Feature Extractor}           & \textit{UNI}                           \\ 
\textbf{Patch Size}                  & 224x224                                \\ 
\textbf{Microns per Pixel (MPP)}     & 1.0                                    \\ 
\textbf{Embedding Dimension} ($d_{\text{model}}$) & 512                         \\ 
\textbf{Transformer Encoder Layers} ($L$) & 2                             \\ 
\textbf{Attention Heads} ($H$)       & 2                                      \\ 
\textbf{Feedforward Dimension} ($d_{\text{ff}}$) & 2048                   \\ 
\textbf{Activation Function}         & ReLU                                   \\ 
\textbf{Dropout Rate}                & 0.15                                   \\ 
\textbf{Layer Norm Epsilon}          & $1 \times 10^{-5}$                     \\ 
\textbf{Loss Function}               & BCEWithLogitsLoss                      \\ 
\textbf{Optimiser}                   & Adam                                   \\ 
\textbf{Learning Rate}               & $1 \times 10^{-4}$                     \\ 
\textbf{Batch Size}                  & 1                                      \\ 
\textbf{Epochs}                      & 200                                    \\ 
\textbf{Automatic Mixed Precision (AMP)}                  & True                                   \\ 
\textbf{GPU}                         & NVIDIA V100 32GB                       \\ 
\bottomrule
 &\\
 &\\
\end{tabular}
\end{table}

\subsection{Experiment Results}

The results underscore the strong utility of the SurGen dataset for developing predictive models in computational pathology. Compared with the previous work\citep{myles2024leveraging}, which achieved a 0.7136 AUROC on the smaller SR386 subset, the higher test AUROC of 0.8273 observed here suggests that SurGen's broader scope and consistently high-quality images may foster more robust model performance. Although additional investigation is necessary to establish whether this improvement stems primarily from the expanded sample size, and greater tumour heterogeneity, these findings emphasise the importance of a large, well-curated dataset for accurate MMR status prediction.

\begin{figure*}[htbp]
\centering
\includegraphics[width=1\textwidth]{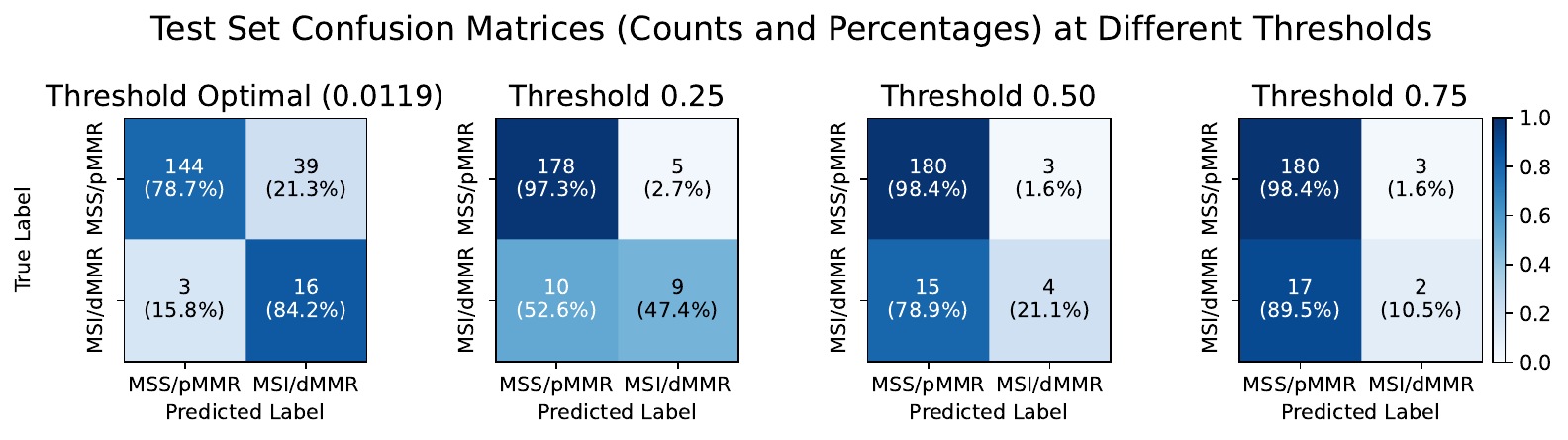}
\caption{Confusion matrices for mismatch repair (MMR) status prediction at various classification thresholds on the test set. The confusion matrices show the classification results for mismatch repair (MMR) status prediction across four different decision thresholds (0.0119, 0.25, 0.50, and 0.75). Threshold 0.0119 represents the point at which 95\% sensitivity on validation set is reached. Each matrix shows the number and percentage of correct and incorrect predictions for the microsatellite-stable/proficient MMR (MSS/pMMR) and microsatellite-instable/deficient MMR (MSI/dMMR) classes.}\label{fig:conf-matrix}
\end{figure*}

The Transformer-based model demonstrated strong performance in predicting MMR status, achieving an AUROC of 0.9297 on the validation set and 0.8273 on the test set. To illustrate how well the model balances sensitivity and specificity, Figure \ref{fig:conf-matrix} shows the confusion matrices at four thresholds, optimal (0.0119), 0.25, 0.50, and 0.75, providing a detailed breakdown of the model's classification performance. These matrices help reveal trade-offs between true positives and false positives under different decision criteria and indicate how threshold selection can be tailored for particular clinical aims. For instance, the 0.0119 threshold achieves 95\% sensitivity on the validation set, which may be important in early-stage colorectal cancer to minimise the chance of missing diseased cases.

\section{Discussion}

In this study, we introduce the SurGen dataset, a comprehensive collection of 1020 H\&E stained WSIs from 843 colorectal cancer cases with detailed genetic and clinical annotations. This dataset addresses the critical need for extensive, high-quality datasets in computational pathology to advance cancer diagnosis and treatment. To demonstrate its utility, we developed a machine learning model capable of predicting mismatch repair (MMR) status from the SurGen dataset, achieving a test AUROC of 0.8273 with no hyperparameter tuning. This performance demonstrates a significant improvement over previous efforts which, despite extensive hyperparameter optimisation on the SR386 subset, achieved an AUROC of only 0.7136 \citep{myles2024leveraging}. This further motivates the need for large and comprehensive WSI datasets to conduct robust and generalisable computational pathology research. The SurGen dataset directly addresses this need by providing a resource that complements existing datasets with its high-resolution WSIs, extensive annotations, and consistent imaging quality.

Unlike many existing datasets, which often suffer from inconsistent image quality which results in users removing subsets of cases \citep{haghighat2022automated, jang2020prediction, lafarge2024image, xu2022spatial, abels2019computational}, SurGen offers over 1000 consistently high-quality WSIs. This ensures researchers can develop and evaluate models on a dataset that reflects real-world high-quality diagnostic conditions.

The SurGen dataset's extensive annotations and high-quality WSIs make it a valuable resource for developing foundational AI models, enabling transfer learning and domain-specific fine-tuning across a wide range of computational pathology tasks.

By providing a robust foundation for algorithm development, the SurGen dataset supports ongoing efforts to personalise cancer diagnosis and treatment strategies at a global scale.

\section{Potential implications}

The SurGen dataset has the potential to impact various areas of cancer research and computational pathology.

In computational pathology, the dataset could serve as a valuable resource for developing and evaluating machine learning algorithms. The diversity of tumour sites and genetic annotations could help in creating more generalisable and robust models. One potential avenue for future research might be to integrate and compare tools such as GrandQC\citep{weng2024grandqc} and others \citep{haghighat2022automated, janowczyk2019histoqc, patil2023efficient} which can aid in performing quality control analysis, and in some instances, precise tissue segmentation. Additional research could be developed with the aim of exploring the clinical tabular data with respect to tumour staging, genetic mutation, and survival analysis. Further work could aim to integrate all of these aspects on top of a computer vision model.

While the dataset originates from a single geographical region, it offers an opportunity to study population-specific cancer characteristics. Comparing SurGen with datasets from other regions may help identify global cancer disparities and inform international research. SurGen, in combination with other international datasets, may offer a broad and comprehensive resource that enhances the generalisability of computational models across diverse populations. This integration can facilitate the development of more robust diagnostic tools that are effective in varied clinical settings, ultimately contributing to a more unified and global approach to cancer diagnosis and treatment. Additionally, leveraging SurGen alongside other datasets can support large-scale studies, enabling researchers to validate findings across different cohorts and improve the reliability of predictive models. Such efforts can drive advancements in personalised medicine, ensuring that computational pathology solutions are both accurate and universally applicable. 

The SurGen dataset has already been adopted in an independent large‑scale benchmark by \citep{vaidya2025molecular}. In that study a survival prediction pipeline was trained and evaluated across seven public cohorts that span multiple tumour types. The resulting top CoXNet~\citep{polsterl2020scikit} and Supervised C‑indices are reproduced in Table~\ref{tab:survival-comparison}. SurGen ranks in the upper half of all cohorts and outperforms several CPTAC datasets of comparable size, thereby providing external evidence of its prognostic signal.

\begin{table}[htbp]
\centering
\caption{Comparison of Survival Prediction Performance Across Datasets. Results are reported as overall survival C-index (mean ± SE) over 5-fold cross-validation. Data is derived and collated from \citep{vaidya2025molecular}.}
\scalebox{0.85}{
\begin{tabular}{lccc}
\toprule
\textbf{Dataset} & \textbf{Patients} & \textbf{CoxNet} & \textbf{Supervised} \\
\midrule
BOEHMK~\citep{boehm2022multimodal}                 & 183  & 0.541 ± 0.013 & 0.575 ± 0.049 \\
\textbf{SURGEN} (ours)                             & 144  & 0.638 ± 0.014 & 0.632 ± 0.022  \\
CPTAC-LUAD~\citep{edwards2015cptac}                & 105  & 0.614 ± 0.032 & 0.576 ± 0.046  \\
CPTAC-HNSC~\citep{edwards2015cptac}                & 102  & 0.631 ± 0.076 & 0.514 ± 0.011  \\
CPTAC-PDAC~\citep{edwards2015cptac}                & 97   & 0.616 ± 0.031 & 0.611 ± 0.042  \\
CPTAC-CCRCC~\citep{edwards2015cptac}               & 94   & 0.675 ± 0.063 & 0.693 ± 0.043  \\
MBC~\citep{bergstrom2024deep, galland2022efficacy} & 75   & 0.550 ± 0.027 & 0.608 ± 0.030 \\

\bottomrule
\end{tabular}
}
\label{tab:survival-comparison}
\end{table}

Ultimately, the SurGen dataset has the potential to accelerate innovations in cancer diagnostics, enhance treatment personalisation, and contribute to reducing the global burden of colorectal cancer.

\section{Availability of Source Code and Requirements}

Source code for data-processing and stratification, background subtraction, feature extraction, model training, and evaluation is available via the project page link below.

\begin{itemize}
\item Project name: SurGen-Dataset
\item Project page: \url{https://github.com/CraigMyles/SurGen-Dataset}
\item Operating system: Ubuntu 20.04 LTS
\item Programming language: Python
\item Other requirements: Pytorch, pylibCZIrw, pandas, NumPy
\item Source code license: GPL-3.0
\item Software Heritage archive: accession swh:1:snp:39dc17fe24087df9ebae119d77d17398aa1ee25a~\citep{myles2025swh}

\end{itemize}

\section{Data Availability}

The dataset supporting this article is available in the European Molecular Biology Laboratory European Bioinformatics Institute (EMBL-EBI) BioImage Archive~\citep{hartley2022bioimage} repository and is available via~\citep{myles2024biostudies} or accessible from within the GitHub README file.

Patch embeddings generated during the preprocessing stages using the UNI foundation model have also been made available to reduce the barrier for entry to researchers wishing to utilise this dataset~\citep{myles2024zenodo}. 

Dome-ML (Data, Optimization, Model and Evaluation in Machine Learning) annotations are available via the DOME registry under accession vuknweu17e~\citep{myles2025domeannotations}. Supporting data is available via the GigaScience database GigaDB~\citep{myles2025gigadb}.

\section{Compute Resource}

In accordance with the recommended minimum documentation for computation time reporting \citep{harris2021understanding}, we have detailed the hardware specifications, computation time, and operating system used during the experiments. 

Feature extraction from WSIs using the UNI foundation model took 2 days, 10 hours, 12 minutes, and 35 seconds on a system equipped with Dual 20-Core Intel Xeon E5-2698 v4 2.2 GHz and a single NVIDIA Tesla V100 32GB GPU. Model training was completed in 3 hours, 2 minutes, and 36 seconds under the same hardware conditions.

\begin{itemize}
\item System: NVIDIA DGX-1
\item Operating System: Ubuntu 20.04 LTS
\item CPU: Dual 20-Core Intel Xeon E5-2698 v4 2.2 GHz
\item GPU: NVIDIA Tesla V100 32GB (Utilised 1 of 8 available)
\item RAM: 512 GB DDR4 RAM
\end{itemize}

\section{Declarations}

\subsection{List of Abbreviations}

AUROC: Area Under the Receiver Operating Characteristic;
BRAF: v-Raf Murine Sarcoma Viral Oncogene Homolog B;
CRC: Colorectal Cancer;
CZI: Carl Zeiss Image (file format);
dMMR: Deficient Mismatch Repair;
FFPE: Formalin-Fixed Paraffin-Embedded;
H\&E: Hematoxylin and Eosin;
IHC: Immunohistochemistry;
KRAS: Kirsten Rat Sarcoma Viral Oncogene Homolog;
MMR: Mismatch Repair;
MSI: Microsatellite Instability;
MSS: Microsatellite Stable;
NGS: Next Generation Sequencing;
NRAS: Neuroblastoma RAS Viral Oncogene Homolog;
PCR: Polymerase Chain Reaction;
TNM: Tumour, Node, Metastasis;
WSI: Whole Slide Image;

\subsection{Ethical Approval}

Ethical approval has been granted by University of St Andrews School of Computer Science Ethics Committee; approval code CS16553. Additionally, Lothian NRS BioResource RTB approval (REC ref – 20/ES/0061 \& 13/ES/0126) has been granted.

\subsection{Consent for Publication}

Not applicable. This manuscript does not contain any individual person's data in a form that would require explicit consent for publication. Comprehensive efforts have been made to ensure patient anonymity. Identifiable information, such as dates of diagnosis, treatment details, and other specifics that could link specimens back to individual patients, have been removed. Furthermore, the dataset has undergone rigorous deidentification processes to aid the prevention re-identification.

\subsection{Competing Interests}

The authors declare that they have no competing interests.

\subsection{Funding}

CM is supported by NHS Lothian. The authors would like to thank NHS Lothian for providing tissue specimens. This work is supported in part by the Industrial Centre for AI Research in Digital Diagnostics (iCAIRD) which is funded by Innovate UK on behalf of UK Research and Innovation (UKRI) (project number 104690).

\subsection{Author's Contributions}

C.M. (C. Myles) led the methodology, investigation, software development, analysis, data curation, and manuscript writing. I.H.U. performed key laboratory work and data acquisition, and contributed to manuscript editing. C.M. (C. Marshall) contributed to data acquisition, governance, and editing. D.H.-B. provided expertise in computational methods, project design, conceptualisation, and manuscript editing. D.J.H. contributed in conceptualisation, clinical insight, and manuscript editing.

\section{Acknowledgements}

The authors would like to thank NHS Lothian for supporting this research and NHS Lothian Biorepository for providing tissue specimens. Special thanks to The Harrison Lab team for their dedicated work in slide processing, digitisation, and genetic and biomarker testing. We also acknowledge the MedTech team in the School of Computer Science at the University of St Andrews for their valuable feedback and support throughout this project.


\begin{thebibliography}{94}
\providecommand{\natexlab}[1]{#1}
\providecommand{\url}[1]{\texttt{#1}}
\expandafter\ifx\csname urlstyle\endcsname\relax
  \providecommand{\doi}[1]{doi: #1}\else
  \providecommand{\doi}{doi: \begingroup \urlstyle{rm}\Url}\fi

\bibitem[Sung et~al.(2021)Sung, Ferlay, Siegel, Laversanne, Soerjomataram, Jemal, and Bray]{sung2021global}
Hyuna Sung, Jacques Ferlay, Rebecca~L Siegel, Mathieu Laversanne, Isabelle Soerjomataram, Ahmedin Jemal, and Freddie Bray.
\newblock Global cancer statistics 2020: Globocan estimates of incidence and mortality worldwide for 36 cancers in 185 countries.
\newblock \emph{CA: a cancer journal for clinicians}, 71\penalty0 (3):\penalty0 209--249, 2021.

\bibitem[Bray et~al.(2024)Bray, Laversanne, Sung, Ferlay, Siegel, Soerjomataram, and Jemal]{bray2024global}
Freddie Bray, Mathieu Laversanne, Hyuna Sung, Jacques Ferlay, Rebecca~L Siegel, Isabelle Soerjomataram, and Ahmedin Jemal.
\newblock Global cancer statistics 2022: Globocan estimates of incidence and mortality worldwide for 36 cancers in 185 countries.
\newblock \emph{CA: a cancer journal for clinicians}, 74\penalty0 (3):\penalty0 229--263, 2024.

\bibitem[Bera et~al.(2019)Bera, Schalper, Rimm, Velcheti, and Madabhushi]{bera2019artificial}
Kaustav Bera, Kurt~A Schalper, David~L Rimm, Vamsidhar Velcheti, and Anant Madabhushi.
\newblock Artificial intelligence in digital pathology—new tools for diagnosis and precision oncology.
\newblock \emph{Nature reviews Clinical oncology}, 16\penalty0 (11):\penalty0 703--715, 2019.

\bibitem[Niazi et~al.(2019)Niazi, Parwani, and Gurcan]{niazi2019digital}
Muhammad Khalid~Khan Niazi, Anil~V Parwani, and Metin~N Gurcan.
\newblock Digital pathology and artificial intelligence.
\newblock \emph{The lancet oncology}, 20\penalty0 (5):\penalty0 e253--e261, 2019.

\bibitem[Abels et~al.(2019)Abels, Pantanowitz, Aeffner, Zarella, Van~der Laak, Bui, Vemuri, Parwani, Gibbs, Agosto-Arroyo, et~al.]{abels2019computational}
Esther Abels, Liron Pantanowitz, Famke Aeffner, Mark~D Zarella, Jeroen Van~der Laak, Marilyn~M Bui, Venkata~NP Vemuri, Anil~V Parwani, Jeff Gibbs, Emmanuel Agosto-Arroyo, et~al.
\newblock Computational pathology definitions, best practices, and recommendations for regulatory guidance: a white paper from the digital pathology association.
\newblock \emph{The Journal of pathology}, 249\penalty0 (3):\penalty0 286--294, 2019.

\bibitem[Litjens et~al.(2018)Litjens, Bandi, Ehteshami~Bejnordi, Geessink, Balkenhol, Bult, Halilovic, Hermsen, Van~de Loo, Vogels, et~al.]{litjens20181399}
Geert Litjens, Peter Bandi, Babak Ehteshami~Bejnordi, Oscar Geessink, Maschenka Balkenhol, Peter Bult, Altuna Halilovic, Meyke Hermsen, Rob Van~de Loo, Rob Vogels, et~al.
\newblock 1399 h\&e-stained sentinel lymph node sections of breast cancer patients: the camelyon dataset.
\newblock \emph{GigaScience}, 7\penalty0 (6):\penalty0 giy065, 2018.

\bibitem[Spanhol et~al.(2015)Spanhol, Oliveira, Petitjean, and Heutte]{spanhol2015dataset}
Fabio~A Spanhol, Luiz~S Oliveira, Caroline Petitjean, and Laurent Heutte.
\newblock A dataset for breast cancer histopathological image classification.
\newblock \emph{Ieee transactions on biomedical engineering}, 63\penalty0 (7):\penalty0 1455--1462, 2015.

\bibitem[Consortium et~al.(2020)]{national2020brca}
National Cancer Institute Clinical Proteomic Tumor~Analysis Consortium et~al.
\newblock The clinical proteomic tumor analysis consortium breast invasive carcinoma collection (cptac-brca).
\newblock \emph{The Cancer Imaging Archive}, 2020.

\bibitem[Da et~al.(2022)Da, Huang, Li, Zuo, Zhang, Liu, Chen, Li, Xu, Hu, et~al.]{da2022digestpath}
Qian Da, Xiaodi Huang, Zhongyu Li, Yanfei Zuo, Chenbin Zhang, Jingxin Liu, Wen Chen, Jiahui Li, Dou Xu, Zhiqiang Hu, et~al.
\newblock Digestpath: A benchmark dataset with challenge review for the pathological detection and segmentation of digestive-system.
\newblock \emph{Medical Image Analysis}, 80:\penalty0 102485, 2022.

\bibitem[Kim et~al.(2023)Kim, Lee, Cho, Kang, Park, Kang, Kim, Choe, Moon, Lee, et~al.]{kim2023paip}
Kyungmo Kim, Kyoungbun Lee, Sungduk Cho, Dong~Un Kang, Seongkeun Park, Yunsook Kang, Hyunjeong Kim, Gheeyoung Choe, Kyung~Chul Moon, Kyu~Sang Lee, et~al.
\newblock Paip 2020: Microsatellite instability prediction in colorectal cancer.
\newblock \emph{Medical Image Analysis}, 89:\penalty0 102886, 2023.

\bibitem[Consortium et~al.(2018)]{national2018luad}
National Cancer Institute Clinical Proteomic Tumor~Analysis Consortium et~al.
\newblock The clinical proteomic tumor analysis consortium lung adenocarcinoma collection (cptac-luad).
\newblock \emph{The Cancer Imaging Archive}, 2018.

\bibitem[Ciardiello et~al.(2022)Ciardiello, Ciardiello, Martini, Napolitano, Tabernero, and Cervantes]{ciardiello2022clinical}
Fortunato Ciardiello, Davide Ciardiello, Giulia Martini, Stefania Napolitano, Josep Tabernero, and Andres Cervantes.
\newblock Clinical management of metastatic colorectal cancer in the era of precision medicine.
\newblock \emph{CA: a cancer journal for clinicians}, 72\penalty0 (4):\penalty0 372--401, 2022.

\bibitem[Weinstein et~al.(2013)Weinstein, Collisson, Mills, Shaw, Ozenberger, Ellrott, Shmulevich, Sander, and Stuart]{weinstein2013cancer}
John~N Weinstein, Eric~A Collisson, Gordon~B Mills, Kenna~R Shaw, Brad~A Ozenberger, Kyle Ellrott, Ilya Shmulevich, Chris Sander, and Joshua~M Stuart.
\newblock The cancer genome atlas pan-cancer analysis project.
\newblock \emph{Nature genetics}, 45\penalty0 (10):\penalty0 1113--1120, 2013.

\bibitem[{National Cancer Institute Clinical Proteomic Tumor Analysis Consortium (CPTAC)}(2020)]{cptac_2020}
{National Cancer Institute Clinical Proteomic Tumor Analysis Consortium (CPTAC)}.
\newblock The clinical proteomic tumor analysis consortium colon adenocarcinoma collection (cptac-coad), 2020.
\newblock URL \url{https://doi.org/10.7937/TCIA.YZWQ-ZZ63}.

\bibitem[Wala et~al.(2024)Wala, de~Bruijn, Coy, Gagne, Chan, Chen, Hoffer, Muhlich, Schultz, Santagata, et~al.]{wala2024integrating}
Jeremiah Wala, Ino de~Bruijn, Shannon Coy, Andreanne Gagne, Sabrina Chan, Yu-An Chen, John Hoffer, Jeremy Muhlich, Nikolaus Schultz, Sandro Santagata, et~al.
\newblock Integrating spatial profiles and cancer genomics to identify immune-infiltrated mismatch repair proficient colorectal cancers.
\newblock \emph{bioRxiv}, pages 2024--09, 2024.

\bibitem[Zhu et~al.(2013)Zhu, Pinsky, Kramer, Prorok, Purdue, Berg, and Gohagan]{zhu2013prostate}
Claire~S Zhu, Paul~F Pinsky, Barnett~S Kramer, Philip~C Prorok, Mark~P Purdue, Christine~D Berg, and John~K Gohagan.
\newblock The prostate, lung, colorectal, and ovarian cancer screening trial and its associated research resource.
\newblock \emph{Journal of the National Cancer Institute}, 105\penalty0 (22):\penalty0 1684--1693, 2013.

\bibitem[Pataki et~al.(2022)Pataki, Olar, Ribli, Pesti, Kontsek, Gy{\"o}ngy{\"o}si, Bilecz, Kov{\'a}cs, Kov{\'a}cs, Kramer, et~al.]{pataki2022huncrc}
B{\'a}lint~{\'A}rmin Pataki, Alex Olar, Dezs{\H{o}} Ribli, Adri{\'a}n Pesti, Endre Kontsek, Benedek Gy{\"o}ngy{\"o}si, {\'A}gnes Bilecz, Tekla Kov{\'a}cs, Krist{\'o}f~Attila Kov{\'a}cs, Zs{\'o}fia Kramer, et~al.
\newblock Huncrc: annotated pathological slides to enhance deep learning applications in colorectal cancer screening.
\newblock \emph{Scientific Data}, 9\penalty0 (1):\penalty0 370, 2022.

\bibitem[Ogino et~al.(2007)Ogino, Kawasaki, Kirkner, Kraft, Loda, and Fuchs]{ogino2007evaluation}
Shuji Ogino, Takako Kawasaki, Gregory~J Kirkner, Peter Kraft, Massimo Loda, and Charles~S Fuchs.
\newblock Evaluation of markers for cpg island methylator phenotype (cimp) in colorectal cancer by a large population-based sample.
\newblock \emph{The Journal of molecular diagnostics}, 9\penalty0 (3):\penalty0 305--314, 2007.

\bibitem[Mirzapoor~Abbasabadi et~al.(2023)Mirzapoor~Abbasabadi, Hamedi~Asl, Rahmani, Shahbadori, Karami, Peymani, Taghizadeh, and Samiee~Rad]{mirzapoor2023kras}
Zohreh Mirzapoor~Abbasabadi, Dariush Hamedi~Asl, Babak Rahmani, Rozhin Shahbadori, Sara Karami, Amir Peymani, Sara Taghizadeh, and Fatemeh Samiee~Rad.
\newblock Kras, nras, braf, and pik3ca mutation rates, clinicopathological association, and their prognostic value in iranian colorectal cancer patients.
\newblock \emph{Journal of clinical laboratory analysis}, 37\penalty0 (5):\penalty0 e24868, 2023.

\bibitem[Guo et~al.(2019)Guo, Wu, Tan, Jin, Sheng, Cai, Liu, and Xu]{guo2019clinicopathologic}
Tian-An Guo, Yu-Chen Wu, Cong Tan, Yu-Tong Jin, Wei-Qi Sheng, San-Jun Cai, Fang-Qi Liu, and Ye~Xu.
\newblock Clinicopathologic features and prognostic value of kras, nras and braf mutations and dna mismatch repair status: a single-center retrospective study of 1,834 chinese patients with stage i--iv colorectal cancer.
\newblock \emph{International journal of cancer}, 145\penalty0 (6):\penalty0 1625--1634, 2019.

\bibitem[De~Roock et~al.(2010)De~Roock, Claes, Bernasconi, De~Schutter, Biesmans, Fountzilas, Kalogeras, Kotoula, Papamichael, Laurent-Puig, et~al.]{de2010effects}
Wendy De~Roock, Bart Claes, David Bernasconi, Jef De~Schutter, Bart Biesmans, George Fountzilas, Konstantine~T Kalogeras, Vassiliki Kotoula, Demetris Papamichael, Pierre Laurent-Puig, et~al.
\newblock Effects of kras, braf, nras, and pik3ca mutations on the efficacy of cetuximab plus chemotherapy in chemotherapy-refractory metastatic colorectal cancer: a retrospective consortium analysis.
\newblock \emph{The lancet oncology}, 11\penalty0 (8):\penalty0 753--762, 2010.

\bibitem[Sclafani et~al.(2020)Sclafani, Wilson, Cunningham, Gonzalez De~Castro, Kalaitzaki, Begum, Wotherspoon, Capdevila, Glimelius, Rosell{\'o}, et~al.]{sclafani2020analysis}
Francesco Sclafani, Sanna~Hulkki Wilson, David Cunningham, David Gonzalez De~Castro, Eleftheria Kalaitzaki, Ruwaida Begum, Andrew Wotherspoon, Jaume Capdevila, Bengt Glimelius, Susana Rosell{\'o}, et~al.
\newblock Analysis of kras, nras, braf, pik3ca and tp53 mutations in a large prospective series of locally advanced rectal cancer patients.
\newblock \emph{International Journal of Cancer}, 146\penalty0 (1):\penalty0 94--102, 2020.

\bibitem[Burotto et~al.(2014)Burotto, Chiou, Lee, and Kohn]{burotto2014mapk}
Mauricio Burotto, Victoria~L Chiou, Jung-Min Lee, and Elise~C Kohn.
\newblock The mapk pathway across different malignancies: a new perspective.
\newblock \emph{Cancer}, 120\penalty0 (22):\penalty0 3446--3456, 2014.

\bibitem[McCain(2013)]{mccain2013mapk}
Jack McCain.
\newblock The mapk (erk) pathway: investigational combinations for the treatment of braf-mutated metastatic melanoma.
\newblock \emph{Pharmacy and Therapeutics}, 38\penalty0 (2):\penalty0 96, 2013.

\bibitem[Li et~al.(2020)Li, Zhao, Yu, and Wei]{li2020braf}
Zi-Nan Li, Lin Zhao, Li-Feng Yu, and Min-Jie Wei.
\newblock Braf and kras mutations in metastatic colorectal cancer: future perspectives for personalized therapy.
\newblock \emph{Gastroenterology report}, 8\penalty0 (3):\penalty0 192--205, 2020.

\bibitem[Boland and Goel(2010)]{boland2010microsatellite}
C~Richard Boland and Ajay Goel.
\newblock Microsatellite instability in colorectal cancer.
\newblock \emph{Gastroenterology}, 138\penalty0 (6):\penalty0 2073--2087, 2010.

\bibitem[Vilar and Gruber(2010)]{vilar2010microsatellite}
Eduardo Vilar and Stephen~B Gruber.
\newblock Microsatellite instability in colorectal cancer—the stable evidence.
\newblock \emph{Nature reviews Clinical oncology}, 7\penalty0 (3):\penalty0 153--162, 2010.

\bibitem[Le et~al.(2017)Le, Durham, Smith, Wang, Bartlett, Aulakh, Lu, Kemberling, Wilt, Luber, et~al.]{le2017mismatch}
Dung~T Le, Jennifer~N Durham, Kellie~N Smith, Hao Wang, Bjarne~R Bartlett, Laveet~K Aulakh, Steve Lu, Holly Kemberling, Cara Wilt, Brandon~S Luber, et~al.
\newblock Mismatch repair deficiency predicts response of solid tumors to pd-1 blockade.
\newblock \emph{Science}, 357\penalty0 (6349):\penalty0 409--413, 2017.

\bibitem[Luchini et~al.(2019)Luchini, Bibeau, Ligtenberg, Singh, Nottegar, Bosse, Miller, Riaz, Douillard, Andre, et~al.]{luchini2019esmo}
C~Luchini, F~Bibeau, MJL Ligtenberg, Navdeep Singh, A~Nottegar, T~Bosse, R~Miller, N~Riaz, J-Y Douillard, F~Andre, et~al.
\newblock Esmo recommendations on microsatellite instability testing for immunotherapy in cancer, and its relationship with pd-1/pd-l1 expression and tumour mutational burden: a systematic review-based approach.
\newblock \emph{Annals of Oncology}, 30\penalty0 (8):\penalty0 1232--1243, 2019.

\bibitem[Lynch et~al.(2009)Lynch, Lynch, Lanspa, Snyder, Lynch, and Boland]{lynch2009review}
Henry~T Lynch, PM~Lynch, SJ~Lanspa, CL~Snyder, JF~Lynch, and CR~Boland.
\newblock Review of the lynch syndrome: history, molecular genetics, screening, differential diagnosis, and medicolegal ramifications.
\newblock \emph{Clinical genetics}, 76\penalty0 (1):\penalty0 1--18, 2009.

\bibitem[Tiwari et~al.(2016)Tiwari, Roy, and Lynch]{tiwari2016lynch}
Ashish~K Tiwari, Hemant~K Roy, and HT~Lynch.
\newblock Lynch syndrome in the 21st century: clinical perspectives.
\newblock \emph{QJM: An International Journal of Medicine}, 109\penalty0 (3):\penalty0 151--158, 2016.

\bibitem[Hampel et~al.(2005)Hampel, Frankel, Martin, Arnold, Khanduja, Kuebler, Nakagawa, Sotamaa, Prior, Westman, et~al.]{hampel2005screening}
Heather Hampel, Wendy~L Frankel, Edward Martin, Mark Arnold, Karamjit Khanduja, Philip Kuebler, Hidewaki Nakagawa, Kaisa Sotamaa, Thomas~W Prior, Judith Westman, et~al.
\newblock Screening for the lynch syndrome (hereditary nonpolyposis colorectal cancer).
\newblock \emph{New England Journal of Medicine}, 352\penalty0 (18):\penalty0 1851--1860, 2005.

\bibitem[Lynch et~al.(2008)Lynch, Lynch, Lynch, and Attard]{lynch2008hereditary}
Henry~T Lynch, Jane~F Lynch, Patrick~M Lynch, and Thomas Attard.
\newblock Hereditary colorectal cancer syndromes: molecular genetics, genetic counseling, diagnosis and management.
\newblock \emph{Familial cancer}, 7:\penalty0 27--39, 2008.

\bibitem[Amin et~al.(2017)Amin, Greene, Edge, Compton, Gershenwald, Brookland, Meyer, Gress, Byrd, and Winchester]{amin2017eighth}
Mahul~B Amin, Frederick~L Greene, Stephen~B Edge, Carolyn~C Compton, Jeffrey~E Gershenwald, Robert~K Brookland, Laura Meyer, Donna~M Gress, David~R Byrd, and David~P Winchester.
\newblock The eighth edition ajcc cancer staging manual: continuing to build a bridge from a population-based to a more “personalized” approach to cancer staging.
\newblock \emph{CA: a cancer journal for clinicians}, 67\penalty0 (2):\penalty0 93--99, 2017.

\bibitem[Dukes(1932)]{dukes1932classification}
Cuthbert~E Dukes.
\newblock The classification of cancer of the rectum.
\newblock \emph{The Journal of Pathology and Bacteriology}, 35\penalty0 (3):\penalty0 323--332, 1932.

\bibitem[against Cancer. Committee~on TNM~Classification(1974)]{international1974tnm}
International~Union against Cancer. Committee~on TNM~Classification.
\newblock \emph{TNM classification of malignant tumours}.
\newblock International Union Against Cancer, 1974.

\bibitem[Haq et~al.(2009)Haq, Schneeweiss, Kalsi, and Arya]{haq2009dukes}
Asif~I Haq, Jenifer Schneeweiss, Vinay Kalsi, and Manit Arya.
\newblock The dukes staging system: a cornerstone in the clinical management of colorectal cancer.
\newblock \emph{The lancet oncology}, 10\penalty0 (11):\penalty0 1128, 2009.

\bibitem[Sobin et~al.(2011)Sobin, Gospodarowicz, and Wittekind]{sobin2011tnm}
Leslie~H Sobin, Mary~K Gospodarowicz, and Christian Wittekind.
\newblock \emph{TNM classification of malignant tumours}.
\newblock John Wiley \& Sons, 2011.

\bibitem[Holub et~al.(2023)Holub, M{\"u}ller, B{\'\i}l, Pireddu, Plass, Prasser, Schl{\"u}nder, Zatloukal, Nenutil, and Br{\'a}zdil]{holub2023privacy}
Petr Holub, Heimo M{\"u}ller, Tom{\'a}{\v{s}} B{\'\i}l, Luca Pireddu, Markus Plass, Fabian Prasser, Irene Schl{\"u}nder, Kurt Zatloukal, Rudolf Nenutil, and Tom{\'a}{\v{s}} Br{\'a}zdil.
\newblock Privacy risks of whole-slide image sharing in digital pathology.
\newblock \emph{Nature Communications}, 14\penalty0 (1):\penalty0 2577, 2023.

\bibitem[Goode et~al.(2013)Goode, Gilbert, Harkes, Jukic, and Satyanarayanan]{goode2013openslide}
Adam Goode, Benjamin Gilbert, Jan Harkes, Drazen Jukic, and Mahadev Satyanarayanan.
\newblock Openslide: A vendor-neutral software foundation for digital pathology.
\newblock \emph{Journal of pathology informatics}, 4\penalty0 (1):\penalty0 27, 2013.

\bibitem[{ZEISS}(2024)]{zeiss_pylibczirw}
{ZEISS}.
\newblock {pylibczirw: A Python wrapper for libCZI}.
\newblock \url{https://github.com/ZEISS/pylibczirw}, 2024.
\newblock Commit ID: 264fcb4ab95274e54433a0054d69f07c402582f4.

\bibitem[Linkert et~al.(2010)Linkert, Rueden, Allan, Burel, Moore, Patterson, Loranger, Moore, Neves, MacDonald, et~al.]{linkert2010metadata}
Melissa Linkert, Curtis~T Rueden, Chris Allan, Jean-Marie Burel, Will Moore, Andrew Patterson, Brian Loranger, Josh Moore, Carlos Neves, Donald MacDonald, et~al.
\newblock Metadata matters: access to image data in the real world.
\newblock \emph{Journal of Cell Biology}, 189\penalty0 (5):\penalty0 777--782, 2010.

\bibitem[Bankhead et~al.(2017)Bankhead, Loughrey, Fern{\'a}ndez, Dombrowski, McArt, Dunne, McQuaid, Gray, Murray, Coleman, et~al.]{bankhead2017qupath}
Peter Bankhead, Maurice~B Loughrey, Jos{\'e}~A Fern{\'a}ndez, Yvonne Dombrowski, Darragh~G McArt, Philip~D Dunne, Stephen McQuaid, Ronan~T Gray, Liam~J Murray, Helen~G Coleman, et~al.
\newblock Qupath: Open source software for digital pathology image analysis.
\newblock \emph{Scientific reports}, 7\penalty0 (1):\penalty0 1--7, 2017.

\bibitem[Schindelin et~al.(2012)Schindelin, Arganda-Carreras, Frise, Kaynig, Longair, Pietzsch, Preibisch, Rueden, Saalfeld, Schmid, et~al.]{schindelin2012fiji}
Johannes Schindelin, Ignacio Arganda-Carreras, Erwin Frise, Verena Kaynig, Mark Longair, Tobias Pietzsch, Stephan Preibisch, Curtis Rueden, Stephan Saalfeld, Benjamin Schmid, et~al.
\newblock Fiji: an open-source platform for biological-image analysis.
\newblock \emph{Nature methods}, 9\penalty0 (7):\penalty0 676--682, 2012.

\bibitem[Schneider et~al.(2012)Schneider, Rasband, and Eliceiri]{schneider2012nih}
Caroline~A Schneider, Wayne~S Rasband, and Kevin~W Eliceiri.
\newblock Nih image to imagej: 25 years of image analysis.
\newblock \emph{Nature methods}, 9\penalty0 (7):\penalty0 671--675, 2012.

\bibitem[Chen et~al.(2024)Chen, Ding, Lu, Williamson, Jaume, Song, Chen, Zhang, Shao, Shaban, et~al.]{chen2024towards}
Richard~J Chen, Tong Ding, Ming~Y Lu, Drew~FK Williamson, Guillaume Jaume, Andrew~H Song, Bowen Chen, Andrew Zhang, Daniel Shao, Muhammad Shaban, et~al.
\newblock Towards a general-purpose foundation model for computational pathology.
\newblock \emph{Nature Medicine}, 30\penalty0 (3):\penalty0 850--862, 2024.

\bibitem[Oquab et~al.(2023)Oquab, Darcet, Moutakanni, Vo, Szafraniec, Khalidov, Fernandez, Haziza, Massa, El-Nouby, et~al.]{oquab2023dinov2}
Maxime Oquab, Timoth{\'e}e Darcet, Th{\'e}o Moutakanni, Huy Vo, Marc Szafraniec, Vasil Khalidov, Pierre Fernandez, Daniel Haziza, Francisco Massa, Alaaeldin El-Nouby, et~al.
\newblock Dinov2: Learning robust visual features without supervision.
\newblock \emph{arXiv preprint arXiv:2304.07193}, 2023.

\bibitem[Oliveira et~al.(2021)Oliveira, Neto, Fraga, Montezuma, Monteiro, Monteiro, Ribeiro, Gon{\c{c}}alves, Pinto, and Cardoso]{oliveira2021cad}
Sara~P Oliveira, Pedro~C Neto, Jo{\~a}o Fraga, Diana Montezuma, Ana Monteiro, Jo{\~a}o Monteiro, Liliana Ribeiro, Sofia Gon{\c{c}}alves, Isabel~M Pinto, and Jaime~S Cardoso.
\newblock Cad systems for colorectal cancer from wsi are still not ready for clinical acceptance.
\newblock \emph{Scientific Reports}, 11\penalty0 (1):\penalty0 14358, 2021.

\bibitem[Cui and Zhang(2021)]{cui2021artificial}
Miao Cui and David~Y Zhang.
\newblock Artificial intelligence and computational pathology.
\newblock \emph{Laboratory Investigation}, 101\penalty0 (4):\penalty0 412--422, 2021.

\bibitem[He et~al.(2016)He, Zhang, Ren, and Sun]{he2016deep}
Kaiming He, Xiangyu Zhang, Shaoqing Ren, and Jian Sun.
\newblock Deep residual learning for image recognition.
\newblock In \emph{Proceedings of the IEEE conference on computer vision and pattern recognition}, pages 770--778, 2016.

\bibitem[Myles et~al.(2024{\natexlab{a}})Myles, Um, Harrison, and Harris-Birtill]{myles2024leveraging}
Craig Myles, In~Hwa Um, David~J Harrison, and David Harris-Birtill.
\newblock Leveraging foundation models for enhanced detection of colorectal cancer biomarkers in small datasets.
\newblock In \emph{Annual Conference on Medical Image Understanding and Analysis}, pages 329--343. Springer, 2024{\natexlab{a}}.

\bibitem[Wang et~al.(2022)Wang, Yang, Zhang, Wang, Zhang, Yang, Huang, and Han]{wang2022transformer}
Xiyue Wang, Sen Yang, Jun Zhang, Minghui Wang, Jing Zhang, Wei Yang, Junzhou Huang, and Xiao Han.
\newblock Transformer-based unsupervised contrastive learning for histopathological image classification.
\newblock \emph{Medical image analysis}, 81:\penalty0 102559, 2022.

\bibitem[Azizi et~al.(2023)Azizi, Culp, Freyberg, Mustafa, Baur, Kornblith, Chen, Tomasev, Mitrovi{\'c}, Strachan, et~al.]{azizi2023robust}
Shekoofeh Azizi, Laura Culp, Jan Freyberg, Basil Mustafa, Sebastien Baur, Simon Kornblith, Ting Chen, Nenad Tomasev, Jovana Mitrovi{\'c}, Patricia Strachan, et~al.
\newblock Robust and data-efficient generalization of self-supervised machine learning for diagnostic imaging.
\newblock \emph{Nature Biomedical Engineering}, 7\penalty0 (6):\penalty0 756--779, 2023.

\bibitem[Chen et~al.(2022)Chen, Chen, Li, Chen, Trister, Krishnan, and Mahmood]{chen2022scaling}
Richard~J Chen, Chengkuan Chen, Yicong Li, Tiffany~Y Chen, Andrew~D Trister, Rahul~G Krishnan, and Faisal Mahmood.
\newblock Scaling vision transformers to gigapixel images via hierarchical self-supervised learning.
\newblock In \emph{Proceedings of the IEEE/CVF Conference on Computer Vision and Pattern Recognition}, pages 16144--16155, 2022.

\bibitem[Kang et~al.(2023)Kang, Song, Park, Yoo, and Pereira]{kang2023benchmarking}
Mingu Kang, Heon Song, Seonwook Park, Donggeun Yoo, and S{\'e}rgio Pereira.
\newblock Benchmarking self-supervised learning on diverse pathology datasets.
\newblock In \emph{Proceedings of the IEEE/CVF Conference on Computer Vision and Pattern Recognition}, pages 3344--3354, 2023.

\bibitem[Filiot et~al.(2023)Filiot, Ghermi, Olivier, Jacob, Fidon, Mac~Kain, Saillard, and Schiratti]{filiot2023scaling}
Alexandre Filiot, Ridouane Ghermi, Antoine Olivier, Paul Jacob, Lucas Fidon, Alice Mac~Kain, Charlie Saillard, and Jean-Baptiste Schiratti.
\newblock Scaling self-supervised learning for histopathology with masked image modeling.
\newblock \emph{medRxiv}, pages 2023--07, 2023.

\bibitem[Lu et~al.(2024)Lu, Chen, Williamson, Chen, Liang, Ding, Jaume, Odintsov, Le, Gerber, et~al.]{lu2024visual}
Ming~Y Lu, Bowen Chen, Drew~FK Williamson, Richard~J Chen, Ivy Liang, Tong Ding, Guillaume Jaume, Igor Odintsov, Long~Phi Le, Georg Gerber, et~al.
\newblock A visual-language foundation model for computational pathology.
\newblock \emph{Nature Medicine}, 30\penalty0 (3):\penalty0 863--874, 2024.

\bibitem[Vorontsov et~al.(2024)Vorontsov, Bozkurt, Casson, Shaikovski, Zelechowski, Severson, Zimmermann, Hall, Tenenholtz, Fusi, et~al.]{vorontsov2024foundation}
Eugene Vorontsov, Alican Bozkurt, Adam Casson, George Shaikovski, Michal Zelechowski, Kristen Severson, Eric Zimmermann, James Hall, Neil Tenenholtz, Nicolo Fusi, et~al.
\newblock A foundation model for clinical-grade computational pathology and rare cancers detection.
\newblock \emph{Nature medicine}, pages 1--12, 2024.

\bibitem[Campanella et~al.(2023)Campanella, Kwan, Fluder, Zeng, Stock, Veremis, Polydorides, Hedvat, Schoenfeld, Vanderbilt, et~al.]{campanella2023computational}
Gabriele Campanella, Ricky Kwan, Eugene Fluder, Jennifer Zeng, Aryeh Stock, Brandon Veremis, Alexandros~D Polydorides, Cyrus Hedvat, Adam Schoenfeld, Chad Vanderbilt, et~al.
\newblock Computational pathology at health system scale--self-supervised foundation models from three billion images.
\newblock \emph{arXiv preprint arXiv:2310.07033}, 2023.

\bibitem[Lai et~al.(2023)Lai, Ahmed, Vijay, Jaroensri, Loo, Vyawahare, Agarwal, Jamil, Matias, Corrado, et~al.]{lai2023domain}
Jeremy Lai, Faruk Ahmed, Supriya Vijay, Tiam Jaroensri, Jessica Loo, Saurabh Vyawahare, Saloni Agarwal, Fayaz Jamil, Yossi Matias, Greg~S Corrado, et~al.
\newblock Domain-specific optimization and diverse evaluation of self-supervised models for histopathology.
\newblock \emph{arXiv preprint arXiv:2310.13259}, 2023.

\bibitem[Hua et~al.(2024)Hua, Yan, Shen, Ma, and Zhang]{hua2024pathoduet}
Shengyi Hua, Fang Yan, Tianle Shen, Lei Ma, and Xiaofan Zhang.
\newblock Pathoduet: Foundation models for pathological slide analysis of h\&e and ihc stains.
\newblock \emph{Medical Image Analysis}, 97:\penalty0 103289, 2024.

\bibitem[Dippel et~al.(2024)Dippel, Feulner, Winterhoff, Milbich, Tietz, Schallenberg, Dernbach, Kunft, Heinke, Eich, et~al.]{dippel2024rudolfv}
Jonas Dippel, Barbara Feulner, Tobias Winterhoff, Timo Milbich, Stephan Tietz, Simon Schallenberg, Gabriel Dernbach, Andreas Kunft, Simon Heinke, Marie-Lisa Eich, et~al.
\newblock Rudolfv: a foundation model by pathologists for pathologists.
\newblock \emph{arXiv preprint arXiv:2401.04079}, 2024.

\bibitem[Aben et~al.(2024)Aben, de~Jong, Gatopoulos, K{\"a}nzig, Karasikov, Lagr{\'e}, Moser, van Doorn, Tang, et~al.]{aben2024towards}
Nanne Aben, Edwin~D de~Jong, Ioannis Gatopoulos, Nicolas K{\"a}nzig, Mikhail Karasikov, Axel Lagr{\'e}, Roman Moser, Joost van Doorn, Fei Tang, et~al.
\newblock Towards large-scale training of pathology foundation models.
\newblock \emph{arXiv preprint arXiv:2404.15217}, 2024.

\bibitem[Juyal et~al.(2024)Juyal, Padigela, Shah, Shenker, Harguindeguy, Liu, Martin, Zhang, Nercessian, Markey, et~al.]{juyal2024pluto}
Dinkar Juyal, Harshith Padigela, Chintan Shah, Daniel Shenker, Natalia Harguindeguy, Yi~Liu, Blake Martin, Yibo Zhang, Michael Nercessian, Miles Markey, et~al.
\newblock Pluto: Pathology-universal transformer.
\newblock \emph{arXiv preprint arXiv:2405.07905}, 2024.

\bibitem[Yang et~al.(2024)Yang, Wei, Liang, Yuan, Gao, Xia, Zhou, Zhang, and Yu]{yang2024foundation}
Zhaochang Yang, Ting Wei, Ying Liang, Xin Yuan, Ruitian Gao, Yujia Xia, Jie Zhou, Yue Zhang, and Zhangsheng Yu.
\newblock A foundation model for generalizable cancer diagnosis and survival prediction from histopathological images.
\newblock \emph{bioRxiv}, pages 2024--05, 2024.

\bibitem[Xu et~al.(2024{\natexlab{a}})Xu, Usuyama, Bagga, Zhang, Rao, Naumann, Wong, Gero, Gonz{\'a}lez, Gu, et~al.]{xu2024whole}
Hanwen Xu, Naoto Usuyama, Jaspreet Bagga, Sheng Zhang, Rajesh Rao, Tristan Naumann, Cliff Wong, Zelalem Gero, Javier Gonz{\'a}lez, Yu~Gu, et~al.
\newblock A whole-slide foundation model for digital pathology from real-world data.
\newblock \emph{Nature}, pages 1--8, 2024{\natexlab{a}}.

\bibitem[Nechaev et~al.(2024)Nechaev, Pchelnikov, and Ivanova]{nechaev2024hibou}
Dmitry Nechaev, Alexey Pchelnikov, and Ekaterina Ivanova.
\newblock Hibou: A family of foundational vision transformers for pathology.
\newblock \emph{arXiv preprint arXiv:2406.05074}, 2024.

\bibitem[Saillard et~al.(2024)Saillard, Jenatton, Llinares-López, Mariet, Cahané, Durand, and Vert]{hoptimus0}
Charlie Saillard, Rodolphe Jenatton, Felipe Llinares-López, Zelda Mariet, David Cahané, Eric Durand, and Jean-Philippe Vert.
\newblock H-optimus-0, 2024.
\newblock URL \url{https://github.com/bioptimus/releases/tree/main/models/h-optimus/v0}.

\bibitem[Xu et~al.(2024{\natexlab{b}})Xu, Wang, Zhou, Ma, Yang, Lin, Wang, Wang, Liang, Han, et~al.]{xu2024multimodal}
Yingxue Xu, Yihui Wang, Fengtao Zhou, Jiabo Ma, Shu Yang, Huangjing Lin, Xin Wang, Jiguang Wang, Li~Liang, Anjia Han, et~al.
\newblock A multimodal knowledge-enhanced whole-slide pathology foundation model.
\newblock \emph{arXiv preprint arXiv:2407.15362}, 2024{\natexlab{b}}.

\bibitem[Zimmermann et~al.(2024)Zimmermann, Vorontsov, Viret, Casson, Zelechowski, Shaikovski, Tenenholtz, Hall, Fuchs, Fusi, et~al.]{zimmermann2024virchow}
Eric Zimmermann, Eugene Vorontsov, Julian Viret, Adam Casson, Michal Zelechowski, George Shaikovski, Neil Tenenholtz, James Hall, Thomas Fuchs, Nicolo Fusi, et~al.
\newblock Virchow 2: Scaling self-supervised mixed magnification models in pathology.
\newblock \emph{arXiv preprint arXiv:2408.00738}, 2024.

\bibitem[Filiot et~al.(2024)Filiot, Jacob, Mac~Kain, and Saillard]{filiot2024phikon}
Alexandre Filiot, Paul Jacob, Alice Mac~Kain, and Charlie Saillard.
\newblock Phikon-v2, a large and public feature extractor for biomarker prediction.
\newblock \emph{arXiv preprint arXiv:2409.09173}, 2024.

\bibitem[Wang et~al.(2024)Wang, Zhao, Marostica, Yuan, Jin, Zhang, Li, Tang, Wang, Li, et~al.]{wang2024pathology}
Xiyue Wang, Junhan Zhao, Eliana Marostica, Wei Yuan, Jietian Jin, Jiayu Zhang, Ruijiang Li, Hongping Tang, Kanran Wang, Yu~Li, et~al.
\newblock A pathology foundation model for cancer diagnosis and prognosis prediction.
\newblock \emph{Nature}, pages 1--9, 2024.

\bibitem[Ding et~al.(2024)Ding, Wagner, Song, Chen, Lu, Zhang, Vaidya, Jaume, Shaban, Kim, et~al.]{ding2024multimodal}
Tong Ding, Sophia~J Wagner, Andrew~H Song, Richard~J Chen, Ming~Y Lu, Andrew Zhang, Anurag~J Vaidya, Guillaume Jaume, Muhammad Shaban, Ahrong Kim, et~al.
\newblock Multimodal whole slide foundation model for pathology.
\newblock \emph{arXiv preprint arXiv:2411.19666}, 2024.

\bibitem[Vaswani et~al.(2017)Vaswani, Shazeer, Parmar, Uszkoreit, Jones, Gomez, Kaiser, and Polosukhin]{vaswani2017attention}
Ashish Vaswani, Noam Shazeer, Niki Parmar, Jakob Uszkoreit, Llion Jones, Aidan~N Gomez, {\L}ukasz Kaiser, and Illia Polosukhin.
\newblock Attention is all you need.
\newblock \emph{Advances in neural information processing systems}, 30, 2017.

\bibitem[Haghighat et~al.(2022)Haghighat, Browning, Sirinukunwattana, Malacrino, Khalid~Alham, Colling, Cui, Rakha, Hamdy, Verrill, et~al.]{haghighat2022automated}
Maryam Haghighat, Lisa Browning, Korsuk Sirinukunwattana, Stefano Malacrino, Nasullah Khalid~Alham, Richard Colling, Ying Cui, Emad Rakha, Freddie~C Hamdy, Clare Verrill, et~al.
\newblock Automated quality assessment of large digitised histology cohorts by artificial intelligence.
\newblock \emph{Scientific Reports}, 12\penalty0 (1):\penalty0 5002, 2022.

\bibitem[Jang et~al.(2020)Jang, Lee, Kang, Song, and Lee]{jang2020prediction}
Hyun-Jong Jang, Ahwon Lee, J~Kang, In~Hye Song, and Sung~Hak Lee.
\newblock Prediction of clinically actionable genetic alterations from colorectal cancer histopathology images using deep learning.
\newblock \emph{World Journal of Gastroenterology}, 26\penalty0 (40):\penalty0 6207, 2020.

\bibitem[Lafarge et~al.(2024)Lafarge, Domingo, Sirinukunwattana, Wood, Samuel, Murray, Richman, Blake, Sebag-Montefiore, Gollins, et~al.]{lafarge2024image}
Maxime~W Lafarge, Enric Domingo, Korsuk Sirinukunwattana, Ruby Wood, Leslie Samuel, Graeme Murray, Susan~D Richman, Andrew Blake, David Sebag-Montefiore, Simon Gollins, et~al.
\newblock Image-based consensus molecular subtyping in rectal cancer biopsies and response to neoadjuvant chemoradiotherapy.
\newblock \emph{NPJ precision oncology}, 8\penalty0 (1):\penalty0 89, 2024.

\bibitem[Xu et~al.(2022)Xu, Cha, Clemenceau, Choi, Lee, Kang, and Hwang]{xu2022spatial}
Hongming Xu, Yoon~Jin Cha, Jean~R Clemenceau, Jinhwan Choi, Sung~Hak Lee, Jeonghyun Kang, and Tae~Hyun Hwang.
\newblock Spatial analysis of tumor-infiltrating lymphocytes in histological sections using deep learning techniques predicts survival in colorectal carcinoma.
\newblock \emph{The Journal of Pathology: Clinical Research}, 8\penalty0 (4):\penalty0 327--339, 2022.

\bibitem[Weng et~al.(2024)Weng, Seper, Pryalukhin, Mairinger, Wickenhauser, Bauer, Glamann, Bl{\"a}ker, Lingscheidt, Hulla, et~al.]{weng2024grandqc}
Zhilong Weng, Alexander Seper, Alexey Pryalukhin, Fabian Mairinger, Claudia Wickenhauser, Marcus Bauer, Lennert Glamann, Hendrik Bl{\"a}ker, Thomas Lingscheidt, Wolfgang Hulla, et~al.
\newblock Grandqc: A comprehensive solution to quality control problem in digital pathology.
\newblock \emph{Nature Communications}, 15\penalty0 (1):\penalty0 1--12, 2024.

\bibitem[Janowczyk et~al.(2019)Janowczyk, Zuo, Gilmore, Feldman, and Madabhushi]{janowczyk2019histoqc}
Andrew Janowczyk, Ren Zuo, Hannah Gilmore, Michael Feldman, and Anant Madabhushi.
\newblock Histoqc: an open-source quality control tool for digital pathology slides.
\newblock \emph{JCO clinical cancer informatics}, 3:\penalty0 1--7, 2019.

\bibitem[Patil et~al.(2023)Patil, Diwakar, Sawant, Kurian, Yadav, Rane, Bameta, and Sethi]{patil2023efficient}
Abhijeet Patil, Harsh Diwakar, Jay Sawant, Nikhil~Cherian Kurian, Subhash Yadav, Swapnil Rane, Tripti Bameta, and Amit Sethi.
\newblock Efficient quality control of whole slide pathology images with human-in-the-loop training.
\newblock \emph{Journal of Pathology Informatics}, 14:\penalty0 100306, 2023.

\bibitem[Vaidya et~al.(2025)Vaidya, Zhang, Jaume, Song, Ding, Wagner, Lu, Doucet, Robertson, Almagro-Perez, et~al.]{vaidya2025molecular}
Anurag Vaidya, Andrew Zhang, Guillaume Jaume, Andrew~H Song, Tong Ding, Sophia~J Wagner, Ming~Y Lu, Paul Doucet, Harry Robertson, Cristina Almagro-Perez, et~al.
\newblock Molecular-driven foundation model for oncologic pathology.
\newblock \emph{arXiv preprint arXiv:2501.16652}, 2025.

\bibitem[P{\"o}lsterl(2020)]{polsterl2020scikit}
Sebastian P{\"o}lsterl.
\newblock scikit-survival: A library for time-to-event analysis built on top of scikit-learn.
\newblock \emph{Journal of Machine Learning Research}, 21\penalty0 (212):\penalty0 1--6, 2020.

\bibitem[Boehm et~al.(2022)Boehm, Aherne, Ellenson, Nikolovski, Alghamdi, V{\'a}zquez-Garc{\'\i}a, Zamarin, Long~Roche, Liu, Patel, et~al.]{boehm2022multimodal}
Kevin~M Boehm, Emily~A Aherne, Lora Ellenson, Ines Nikolovski, Mohammed Alghamdi, Ignacio V{\'a}zquez-Garc{\'\i}a, Dmitriy Zamarin, Kara Long~Roche, Ying Liu, Druv Patel, et~al.
\newblock Multimodal data integration using machine learning improves risk stratification of high-grade serous ovarian cancer.
\newblock \emph{Nature cancer}, 3\penalty0 (6):\penalty0 723--733, 2022.

\bibitem[Edwards et~al.(2015)Edwards, Oberti, Thangudu, Cai, McGarvey, Jacob, Madhavan, and Ketchum]{edwards2015cptac}
Nathan~J Edwards, Mauricio Oberti, Ratna~R Thangudu, Shuang Cai, Peter~B McGarvey, Shine Jacob, Subha Madhavan, and Karen~A Ketchum.
\newblock The cptac data portal: a resource for cancer proteomics research.
\newblock \emph{Journal of proteome research}, 14\penalty0 (6):\penalty0 2707--2713, 2015.

\bibitem[Bergstrom et~al.(2024)Bergstrom, Abbasi, D{\'\i}az-Gay, Galland, Ladoire, Lippman, and Alexandrov]{bergstrom2024deep}
Erik~N Bergstrom, Ammal Abbasi, Marcos D{\'\i}az-Gay, Lo{\"\i}ck Galland, Sylvain Ladoire, Scott~M Lippman, and Ludmil~B Alexandrov.
\newblock Deep learning artificial intelligence predicts homologous recombination deficiency and platinum response from histologic slides.
\newblock \emph{Journal of Clinical Oncology}, 42\penalty0 (30):\penalty0 3550--3560, 2024.

\bibitem[Galland et~al.(2022)Galland, Ballot, Mananet, Boidot, Lecuelle, Albuisson, Arnould, Desmoulins, Mayeur, Kaderbhai, et~al.]{galland2022efficacy}
Lo{\"\i}ck Galland, Elise Ballot, Hugo Mananet, Romain Boidot, Julie Lecuelle, Juliette Albuisson, Laurent Arnould, Isabelle Desmoulins, Didier Mayeur, Cour{\`e}che Kaderbhai, et~al.
\newblock Efficacy of platinum-based chemotherapy in metastatic breast cancer and hrd biomarkers: utility of exome sequencing.
\newblock \emph{NPJ breast cancer}, 8\penalty0 (1):\penalty0 28, 2022.

\bibitem[Myles(2025)]{myles2025swh}
Craig Myles.
\newblock Surgen-dataset (version 1) [computer software].
\newblock \url{https://archive.softwareheritage.org/swh:1:snp:39dc17fe24087df9ebae119d77d17398aa1ee25a}, 2025.
\newblock Software Heritage snapshot.

\bibitem[Hartley et~al.(2022)Hartley, Kleywegt, Patwardhan, Sarkans, Swedlow, and Brazma]{hartley2022bioimage}
Matthew Hartley, Gerard~J Kleywegt, Ardan Patwardhan, Ugis Sarkans, Jason~R Swedlow, and Alvis Brazma.
\newblock The bioimage archive--building a home for life-sciences microscopy data.
\newblock \emph{Journal of Molecular Biology}, 434\penalty0 (11):\penalty0 167505, 2022.

\bibitem[Myles et~al.(2024{\natexlab{b}})Myles, Um, Marshall, Harris-Birtill, and Harrison]{myles2024biostudies}
Craig Myles, In~Hwa Um, Craig Marshall, David Harris-Birtill, and David~J Harrison.
\newblock Surgen: 1020 h\&e-stained whole slide images with survival and genetic markers.
\newblock \url{https://doi.org/10.6019/S-BIAD1285}, 2024{\natexlab{b}}.

\bibitem[Myles et~al.(2024{\natexlab{c}})Myles, Um, Marshall, Harris-Birtill, and Harrison]{myles2024zenodo}
Craig Myles, In~Hwa Um, Craig Marshall, David Harris-Birtill, and David~J Harrison.
\newblock Patch-level uni feature embeddings from colorectal cancer whole slide images (wsis) in the surgen dataset, December 2024{\natexlab{c}}.
\newblock URL \url{https://doi.org/10.5281/zenodo.14047723}.

\bibitem[myl(2025)]{myles2025domeannotations}
Dome-ml annotations for ``surgen: 1020 h\&e-stained whole slide images with survival and genetic markers'' [dome registry].
\newblock \url{https://registry.dome-ml.org/review/vuknweu17e}, 2025.
\newblock DOME registry accession vuknweu17e.

\bibitem[Myles et~al.(2025)Myles, Um, Marshall, Harris-Birtill, and Harrison]{myles2025gigadb}
Craig Myles, In~Hwa Um, Craig Marshall, David Harris-Birtill, and David~J Harrison.
\newblock Supporting data for ``surgen: 1020 h\&e-stained whole slide images with survival and genetic markers''.
\newblock \url{https://doi.org/10.5524/102725}, 2025.
\newblock Dataset, GigaScience Database.

\bibitem[Harris-Birtill and Harris-Birtill(2021)]{harris2021understanding}
David Harris-Birtill and Rose Harris-Birtill.
\newblock Understanding computation time: a critical discussion of time as a computational performance metric.
\newblock In \emph{Time in Variance}, pages 220--248. Brill, 2021.

\end{thebibliography}

\end{document}